\documentclass[12pt]{article}
\usepackage{amsmath}
\usepackage{graphicx,psfrag,epsf,color,subfigure}
\usepackage{enumerate}
\usepackage{natbib}

\usepackage[unicode=true,pdfusetitle,
   bookmarks=true, 
   bookmarksnumbered=false, 
   bookmarksopen=false,
   breaklinks=false, 
   pdfborder={0 0 1}, 
   backref=false, 
   colorlinks=false]{hyperref}

\usepackage{amssymb}
\usepackage{multirow}
\usepackage{array}
\usepackage{bm}
\usepackage{booktabs}
\usepackage{float}
\usepackage[font=small,labelfont=bf]{caption}
\setcitestyle{citesep={;}}
\usepackage[total={6.4in,8.6in}]{geometry}
\usepackage[utf8]{inputenc} 
\usepackage[T1]{fontenc}    
\usepackage{hyperref}       
\usepackage{url}            
\usepackage{booktabs}       
\usepackage{amsfonts}       
\usepackage{nicefrac}       
\usepackage{microtype}      
\usepackage{xcolor}         

\usepackage{amsmath,amsfonts,bm}

\def\eqref#1{equation~\ref{#1}}

\def\floor#1{\lfloor #1 \rfloor}
\def\1{\bm{1}}

\DeclareMathAlphabet{\mathsfit}{\encodingdefault}{\sfdefault}{m}{sl}
\SetMathAlphabet{\mathsfit}{bold}{\encodingdefault}{\sfdefault}{bx}{n}

\newcommand{\Var}{\mathrm{Var}}

\DeclareMathOperator*{\argmax}{arg\,max}
\DeclareMathOperator*{\argmin}{arg\,min}

\usepackage{algorithm2e} 
\usepackage{graphicx,psfrag,epsf,color}
\usepackage{amssymb}
\usepackage{epsfig, graphicx, amsmath}
\usepackage{subfigure}

\newcommand{\Mean}{{E}}

\newcommand{\prob}{{\mbox{Pr}}}

\usepackage{authblk}

\newtheorem{lemma}{Lemma}[section]
\newtheorem{theorem}{Theorem}

 \newcommand*\samethanks[1][\value{footnote}]{\footnotemark[#1]}
 
\begin{document}

\def\spacingset#1{\renewcommand{\baselinestretch}%
{#1}\small\normalsize} \spacingset{1}
%
%
\title{\bf Deep Jump Learning for Off-Policy Evaluation in Continuous Treatment Settings}
\author[1]{Hengrui Cai  \thanks{Equal contribution.} \thanks{hcai5@ncsu.edu} }
\author[2]{Chengchun Shi  \samethanks[1] \thanks{C.Shi7@lse.ac.uk} }
\author[1]{Rui Song\thanks{rsong@ncsu.edu}}
\author[1]{Wenbin Lu\thanks{wlu4@ncsu.edu} \thanks{This paper is accepted at the 35th Conference on Neural Information Processing Systems (NeurIPS 2021). The authors are grateful to the anonymous reviewers for valuable comments and suggestions. Chengchun Shi's research is partially supported by the London School of Economics and Political Science research support fund. Rui Song's research is partially supported by a grant from the National Science Foundation DMS-2113637.}} 
\affil[1]{Department of Statistics, North Carolina State University}
\affil[2]{Department of Statistics, London School of Economics and Political Science}
 \date{}
 \maketitle 

\baselineskip=21pt

%



\begin{abstract}  
We consider off-policy evaluation (OPE) in continuous treatment settings, such as personalized dose-finding. In OPE, one aims to estimate the mean outcome under a new treatment decision rule using historical data generated by a different decision rule. Most existing works on OPE focus on discrete treatment settings. To handle continuous treatments, we develop a novel estimation method for OPE using deep jump learning. The key ingredient of our method lies in adaptively discretizing the treatment space using deep discretization, by leveraging deep learning and multi-scale change point detection. This allows us to apply existing OPE methods in discrete treatments to handle continuous treatments. Our method is further justified by theoretical results, simulations, and a real application to Warfarin Dosing.
\end{abstract}

\section{Introduction}\label{intro}
Individualization proposes to leverage omni-channel data to meet individual needs. Individualized decision making plays a vital role in a wide variety of applications. Examples include individualized treatment regime in precision medicine \citep{qian2011performance,chakraborty2013statistical,wang2018quantile,chen2020representation}, customized pricing strategy in economics \citep{qiang2016dynamic, turvey2017optimal}, personalized recommendation system in marketing \citep{mcinerney2018explore}, etc. Prior to adopting any decision rule in practice, it is crucial to know the impact of implementing such a policy. In medical and public-policy domains, it is risky to apply a treatment decision rule or policy online to estimate its mean outcome \citep[see, e.g.,][]{murphy2001marginal,hirano2003efficient,li2011unbiased}. Off-policy evaluation (OPE) thus attracts a lot of attention by estimating the mean outcome under a new decision rule (or policy), i.e., the value function, using the offline data generated by a different decision rule. 

 Despite the popularity of developing OPE methods with finitely many treatment (or action) options {\citep[see e.g.,][]{dudik2011doubly, dudik2014doubly, wang2012evaluation, zhang2012robust, zhang2013robust,luedtke2016statistical,jiang2016doubly,swaminathan2017off, wang2017optimal,farajtabar2018more,cai2020validation,wu2020resampling,su2020doubly,kallus2020double,shi2020breaking,shi2021statistical}}, less attention has been paid to the continuous treatment setting, such as personalized dose finding {\citep{chen2016personalized,zhou2018parsimonious,zhu2020kernel,zhu2020causal}}, dynamic pricing \citep{den2020discontinuous}, {and contextual bandits \citep{chernozhukov2019semi}.} 
Recently, a few OPE methods have been proposed to handle continuous treatments  {\citep{kallus2018policy,krishnamurthy2019contextual,sondhi2020balanced,colangelo2020double,singh2020kernel,su2020adaptive,kallus2020doubly}}. All these methods rely on the use of a kernel function to extend the inverse probability weighting (IPW) or doubly robust (DR) approaches developed in discrete treatment domains \citep[see e.g.,][]{dudik2011doubly}. They suffer from two limitations. \textbf{First, the validity of these methods requires the mean outcome to be a smooth function over the treatment space}. This assumption could be violated in applications such as dynamic pricing, where the expected demand for a product has jump discontinuities as a function of the charged price \citep{den2020discontinuous}. {Specifically, a product could attract a new segment of customers if the seller lowers the price below a certain threshold. Thus, there will be a sudden increase in demand by a small price reduction, yielding a discontinuous demand function.} 
\textbf{Second, these kernel-based methods typically use a single bandwidth parameter}. This is sub-optimal in cases where the second-order derivative of the conditional mean function has an abrupt change in the treatment space; see Section \ref{sec:toy} for details. Addressing these limitations requires the development of new policy evaluation tools and theory. 

%
  

Our contributions are summarized as follows.
Methodologically, we develop a deep jump learning (DJL) method by integrating {deep learning} \citep{lecun2015deep}, {multi-scale change point detection} \citep[see e.g.,][for an overview]{niu2016multiple}, and the {doubly-robust} value estimators in discrete domains. \textbf{Our method does not require kernel bandwidth selection. It does not suffer from the limitations of kernel-based methods.} The key ingredient of our method lies in adaptively discretizing the treatment space using deep discretization. This allows us to apply the IPW or DR methods to derive the value estimator. {The discretization addresses the first limitation of kernel-based methods, allowing us to handle discontinuous value functions. The adaptivity addresses the second limitation of kernel-based methods. Specifically, it guarantees the optimality of the proposed method in cases where the second-order derivative of the conditional mean function has an abrupt change in the treatment space.}
Theoretically, we derive the convergence rate of the value estimator under DJL for any policy of interest, allowing the conditional mean outcome to be either a continuous or piecewise function of the treatment; see Theorems \ref{thm1} and \ref{thm2} for details. \textbf{Under the piecewise model assumption, the rate of convergence is faster than kernel-based OPE methods}. Under the continuous model assumption, kernel-based estimators converge at a slower rate when the bandwidth undersmoothes or oversmoothes the data. Proofs of our theorems rely on establishing the uniform rate of convergence of deep learning estimators; see Lemma \ref{lemma0} in the supplementary article. We expect this result to be of general interest in contributing to the line of work on developing theories for deep learning methods \citep[see e.g.,][]{imaizumi2019deep,schmidt2020nonparametric,farrell2021deep}. 
Empirically, we show the proposed deep jump learning outperforms existing state-of-the-art OPE methods in both simulations and a real data application to warfarin dosing.  


\section{Preliminaries}\label{sec:pre} 
We first formulate the OPE problem in continuous domains. We next review some related literature on the DR value estimator, kernel based evaluation methods, and multi-scale change point detection.
\subsection{Problem Formulation}\label{sec:prob}
The observed offline datasets can be summarized into  $\{(X_i,A_i,Y_i)\}_{1\le i\le n}$ where $O_i=(X_i,A_i,Y_i)$ denotes the feature-treatment-outcome triplet for the $i$th individual and $n$ denotes the total sample size. We assume these data triplets are independent copies of the population variables $(X,A,Y)$. Let $\mathcal{X}\in \mathbb{R}^p$ and $\mathcal{A}$ denote the $p$ dimensional feature and treatment (or action) space, respectively. We focus on the setting where $\mathcal{A}$ is one-dimensional, as in personalized dose finding and dynamic pricing.  
A decision rule or policy $\pi: \mathcal{X} \to \mathcal{A}$ determines the treatment to be assigned given the observed feature. We use $b$ to denote the propensity score, also known as the behavior policy, that generates the observed data. Specifically, $b(\bullet|x)$ denotes the probability density function of $A$ given $X=x$. 
Define the expected outcome function conditional on the feature-treatment pair as 
\begin{eqnarray*}
Q(x, a) = E(Y|X=x,A=a).
\end{eqnarray*} 
As standard in the OPE and the causal inference literature \citep[see e.g.,][]{chen2016personalized,kallus2018policy}, we assume the stable unit treatment value assumption, no unmeasured confounders assumption, and the positivity assumption are satisfied.  
The positivity assumption requires $b$ to be uniformly bounded away from zero. The latter two assumptions are automatically satisfied in randomized studies. 
These three assumptions guarantee that a policy's value is estimable from the observed data. 
Specifically, for a target policy $\pi$ of interest, its value can be represented by 
\begin{eqnarray*}
V(\pi) = E [Q\{X, \pi(X)\}].
\end{eqnarray*}  
Our goal is to estimate the value $ V(\pi) $ based on the observed data.
 
\subsection{Doubly Robust Estimator and Kernel-Based Evaluation}
For discrete treatments, \citet{dudik2011doubly} proposed a DR estimator of $V(\pi)$ by
\begin{eqnarray}\label{eqn:DR}
\frac{1}{n}\sum_{i=1}^n \psi(O_i,\pi,\widehat{Q},\widehat{b})=\frac{1}{n}\sum_{i=1}^n \left[\widehat{Q}\{X_i,\pi(X_i)\}+\frac{\mathbb{I}\{A_i=\pi(X_i)\}}{\widehat{b}(A_i|X_i)}\{Y_i-\widehat{Q}(X_i,A_i)\}\right],
\end{eqnarray} 
where $\mathbb{I}(\bullet)$ denotes the indicator function, $\widehat{Q}$ and $\widehat{b}(a|x)$ denote some estimators for the conditional mean function $Q$ and the propensity score $b$, respectively. The second term 
inside the bracket corresponds to an augmentation term. Its expectation equals zero when $\widehat{Q}=Q$. The purpose of adding this term is to offer additional protection against potential model misspecification of $Q$. Such an estimator is doubly-robust in the sense that its consistency relies on either the estimation model of ${Q}$ or ${b}$ to be correctly specified. {It can be semi-parametrically efficient whereas the importance sampling or direct method are not.} By setting $\widehat{Q}=0$, \eqref{eqn:DR} reduces to the IPW estimator. 

In continuous treatment domains, the indicator function $\mathbb{I}\{A_i=\pi(X_i)\}$ equals zero almost surely. Consequently, naively applying \eqref{eqn:DR} yields a plug-in estimator $\sum_{i=1}^n \widehat{Q}\{X_i,\pi(X_i)\}/n$. To address this concern, the kernel-based methods proposed to replace the indicator function in \eqref{eqn:DR} with a kernel function $K[\{A_i-\pi(X_i)\}/h]$, i.e.,
\begin{eqnarray}\label{eqn:kernel}
	\frac{1}{n}\sum_{i=1}^n \psi_h(O_i,\pi,\widehat{Q},\widehat{b})=\frac{1}{n}\sum_{i=1}^n \left[\widehat{Q}\{X_i,\pi(X_i)\}+\frac{K\{{A_i-\pi(X_i)\over h}\}}{\widehat{b}(A_i|X_i)}\{Y_i-\widehat{Q}(X_i,A_i)\}\right].
\end{eqnarray}
Here, the bandwidth $h$ represents a trade-off. The variance of the resulting value estimator decays with $h$. Yet, its bias increases with $h$. More specifically, it follows from Theorem 1 of \cite{kallus2018policy} that the leading term of the bias is equal to
\begin{eqnarray}\label{eqn:bias}
h^2\frac{\int u^2 K(u)du}{2} \Mean \left\{\left.\frac{\partial^2 Q(X,a)}{\partial a^2}\right|_{a=\pi(X)}\right\}.
\end{eqnarray}
To ensure the term in \eqref{eqn:bias} decays to zero as $h$ goes to 0, it requires the expected second derivative of the function $Q(x,a)$ exists, and thus $Q(x,a)$ needs to be a smooth function of $a$. However, as commented in the introduction, this assumption could be violated in certain applications.

%
\subsection{Multi-Scale Change Point Detection}

To adaptively discretize the treatment space, we leverage ideas from multi-scale change point detection literature. 
The change point analysis considers an ordered sequence of data, $Y_{1:n} = \{Y_1, \cdots, Y_n\}$, with unknown change point locations, $\tau = \{\tau_1, \cdots, \tau_K\}$ for some unknown integer $K$. Here, $\tau_i$ is an integer between 1 and $n-1$ inclusive, and satisfies $\tau_i <\tau_j $ for $ i < j$. These change points split the data into $K+1$ segments{. Assume there are sufficiently many data points lying within each segment such that the expected reward can be consistently estimated}. Within each segment, the expected outcome is a constant function; see the left panel of Figure \ref{fig:toy} for details. A number of methods have been proposed on estimating change points \citep[see for example,][and the references therein]{boysen2009consistencies, killick2012optimal, Frick2014, Piotr2014, fryzlewicz2020narrowest}, by minimizing a penalized objective function:
\begin{equation*}
\argmin_{\tau } \Bigg({1\over n} \sum_{i=1}^{K+1} \Big[\mathcal{C}\{Y_{(\tau_{i-1}+1):\tau_i}\} \Big] + \gamma_n K \Bigg),
\end{equation*}
where $\mathcal{C}$ is a cost function that measures the goodness-of-the-fit of the constant function within each segment and $\gamma_n K$ penalizes the number of change points. We remark that all
the above cited works focused on either models without features or linear models. Our proposal goes beyond these
works in that we consider models with features and use deep neural networks (DNN) to
capture the complex relationship between the outcome and features.

\section{Deep Jump Learning}\label{model}
 In Section \ref{sec:toy}, we use a toy example to demonstrate the limitation of kernel-based methods. We present the main idea of our algorithm in Section \ref{sec:mainidea}. Details are given in Section \ref{sec:algo}. {For simplicity, we set the action space $ \mathcal{A}=[0,1]$. Define a discretization $\mathcal{D}$ for the treatment space $\mathcal{A}$ as a set of mutually disjoint intervals $\{[\tau_0,\tau_1),[\tau_1,\tau_2),\dots,[\tau_{K-1},\tau_K]\}$ for some  $0=\tau_0<\tau_1<\tau_2<\cdots<\tau_{K-1}< \tau_K=1$ and some integer $K\ge 1$. The union of these intervals covers $\mathcal{A}$. We use $J(\mathcal{D})$ to denote the set of change point locations, i.e., $\{\tau_1,\cdots,\tau_{K-1}\}$. 
We use $|\mathcal{D}|$ to denote the number of intervals in $\mathcal{D}$ and $|\mathcal{I}|$ to denote the length of any interval $\mathcal{I}$. }


\subsection{Toy Example}\label{sec:toy}

\begin{figure} 
 \centering
 \begin{subfigure}{}
 \centering
 		\includegraphics[width=0.3\linewidth,height=0.2\linewidth]{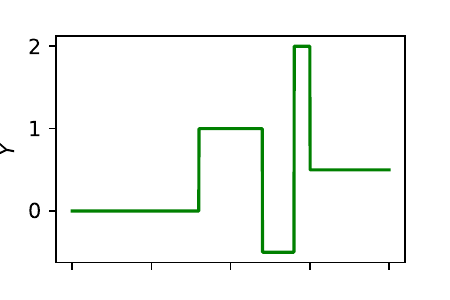}
 \end{subfigure}
\begin{subfigure}{}
	\centering
	\includegraphics[width=0.32\linewidth,height=0.23\linewidth]{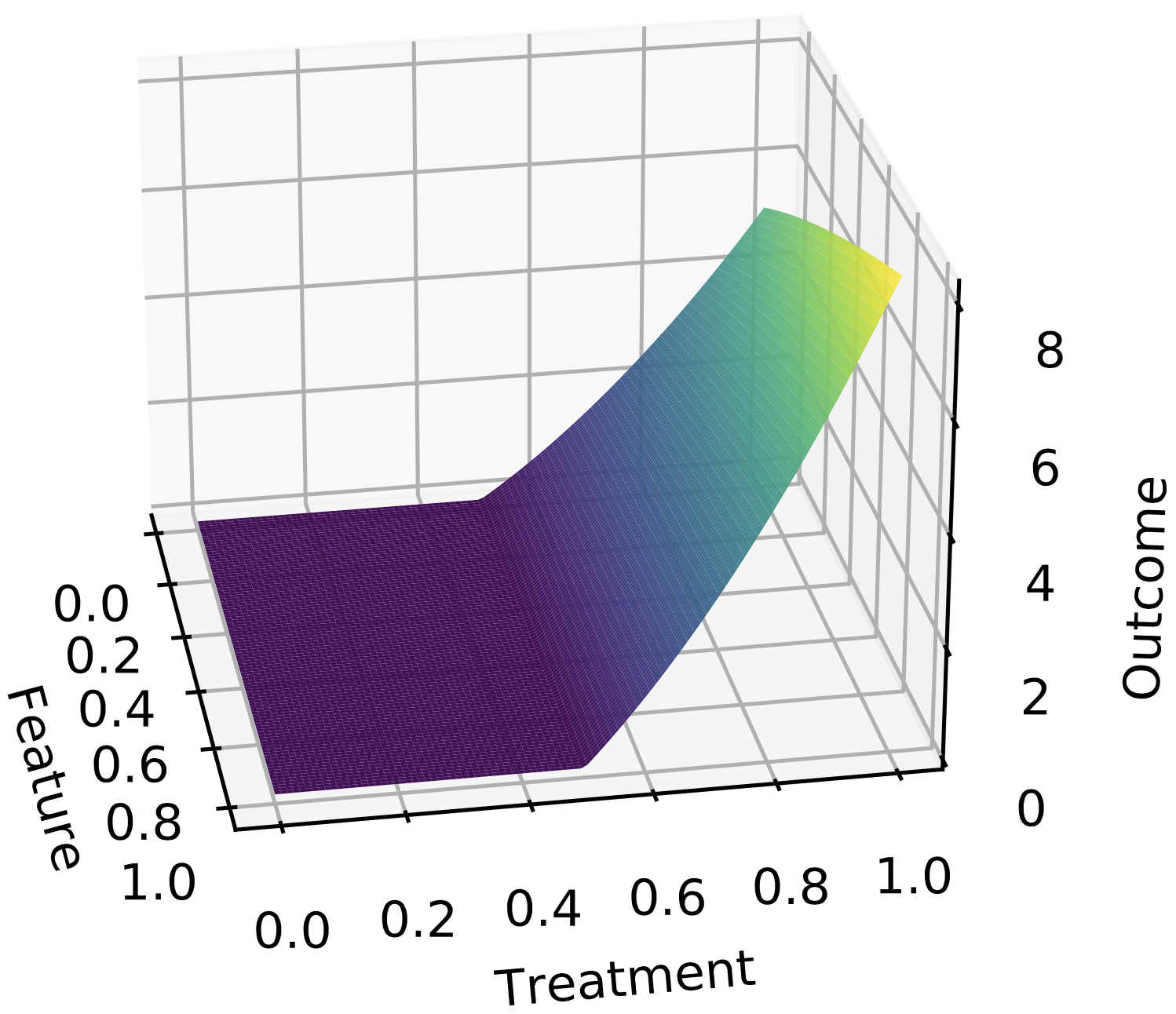}
\end{subfigure}
\begin{subfigure}{}
	\centering
	\includegraphics[width=0.32\linewidth,height=0.23\linewidth]{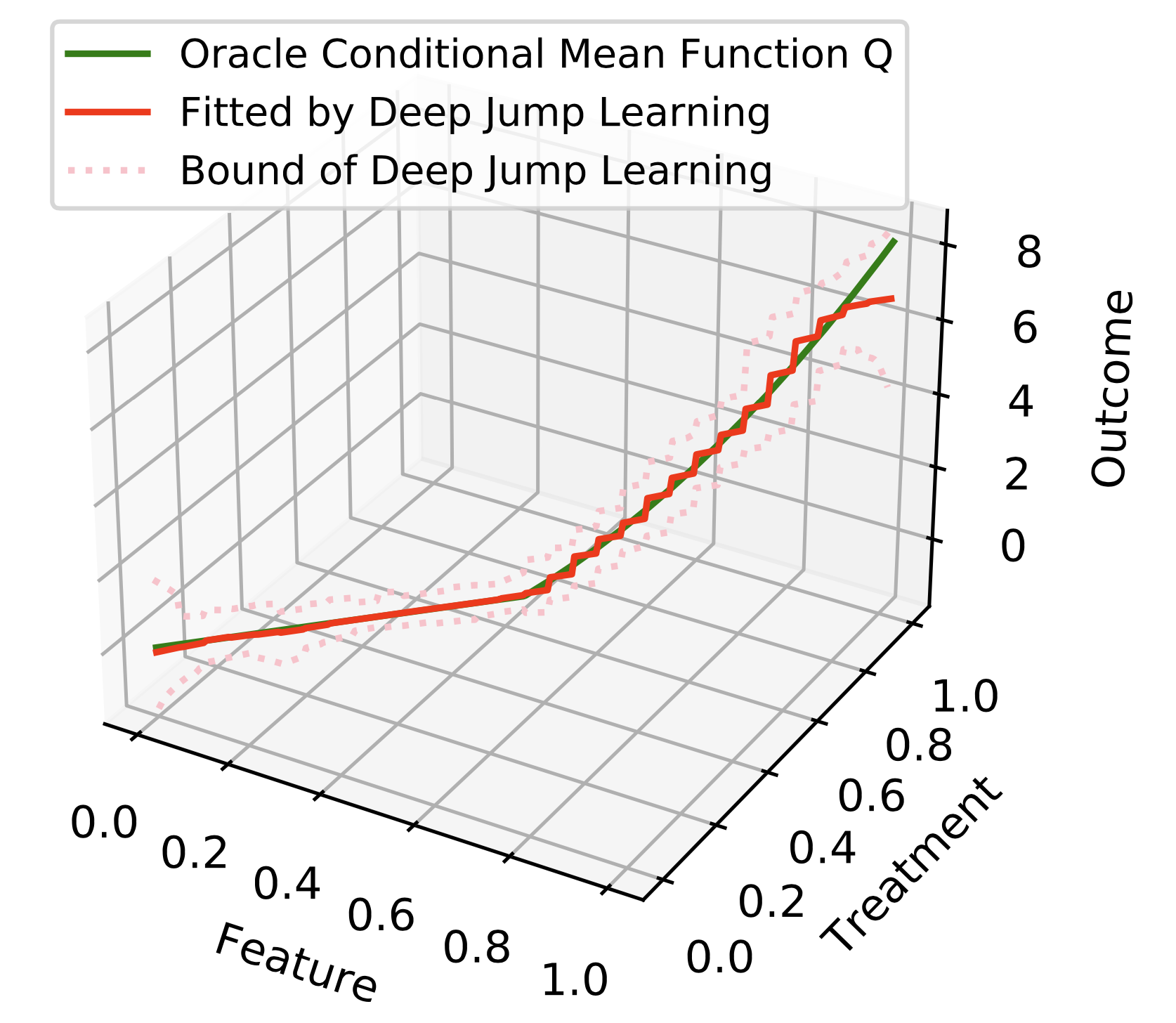}
\end{subfigure}
	\caption{Left panel: example of piece-wise constant function. Middle panel: the oracle conditional mean function $Q$ on the feature-treatment space for the toy example. Right panel: the green curve presents the oracle $Q\{x,\pi(x)\}$ under target policy $\pi(x)=x$ in the toy example; and the red curve is the fitted mean value by DJL and the pink dash line corresponds to the 95\% confidence bound.}
	\label{fig:toy}
\end{figure}

As discussed in the introduction, existing kernel-based methods use a single bandwidth to construct the value estimator. 
Ideally, the bandwidth $h$ in the kernel $K[\{A_i-\pi(X_i)\}/h]$ shall vary with $\pi(X_i)$ to improve the accuracy of the value estimator. To elaborate this, 
consider the function $Q(x, a) = 10 \max(a^2-0.25,0) \log(x+2) $ for any $x, a\in [0,1]$. By definition, $Q(x, a)$ is smooth over the entire feature-treatment space. However, it has different patterns when the treatment belongs to different intervals. Specifically, for $a\in [0,0.5]$, $Q(x,a)$ is constant as a function of $a$. For $a\in (0.5,1]$, $Q(x,a)$ depends quadratically in $a$. See the middle panel of Figure \ref{fig:toy} for details. 

Consider the target policy $\pi(x)=x$. We decompose the value $V(\pi)$ into $V^{(1)}(\pi)+V^{(2)}(\pi)$ where 
\begin{eqnarray*} 
	V^{(1)}(\pi)=\Mean[ Q\{X,\pi(X)\}\mathbb{I}\{\pi(X)\le 0.5\}], \text{ and } V^{(2)}(\pi)=\Mean[ Q\{X,\pi(X)\}\mathbb{I}\{\pi(X)> 0.5\}].
	\end{eqnarray*} 
Similarly, denote the corresponding kernel-based value estimators by {
\begin{eqnarray*} 
	\widehat{V}^{(1)}(\pi;h)=\frac{1}{n}\sum_{i=1}^n  [\psi_h \mathbb{I}\{\pi(X_i)\le 0.5\} ],\text{ and } 
	\widehat{V}^{(2)}(\pi;h)=\frac{1}{n}\sum_{i=1}^n [ \psi_h \mathbb{I}\{\pi(X_i)> 0.5\} ],
\end{eqnarray*} 
where $\psi_h:=\psi_h(O_i,\pi,\widehat{Q},\widehat{b})$} is defined in \eqref{eqn:kernel}. 
Since $Q(x,a)$ is a constant function of $a\in [0,0.5]$, its second-order derivative $\partial^2 Q(x,a)/\partial a^2$ equals zero. In view of \eqref{eqn:bias}, when $\pi(x)\le 0.5$, the bias of $\widehat{V}^{(1)}(\pi;h)$ will be small even with a sufficiently large $h$. As such, a large $h$ is preferred to reduce the variance of $\widehat{V}^{(1)}(\pi;h)$. When $\pi(x)>0.5$, a small $h$ is preferred to reduce the bias of $\widehat{V}^{(2)}(\pi;h)$. 
A simulation study is provided to demonstrate the drawback of the kernel-based methods. 
Specifically, we set $X,A\sim \textrm{Uniform}[0,1]$ and generate $Y|X,A\sim N\{Q(X,A),1\}$. 
We apply the kernel-based methods with a Gaussian  kernel to estimate $V^{(1)}(\pi)$ and $V^{(2)}(\pi)$ with the sample size $n=300$ over 100 replications. See Table \ref{table:toy} for details of the absolute error and standard deviation of $\widehat{V}^{(1)}(\pi;h)$ and $\widehat{V}^{(2)}(\pi;h)$ 
with two different bandwidths $h =0.4$ and $1$. It can be seen that \textbf{due to the use of a single bandwidth, the kernel-based estimator suffers from either a large absolute error or a large variance}.

\begin{table}

\centering
	\caption{The absolute error and the standard deviation (in parentheses) of the estimated values for $V^{(1)}$ and $V^{(2)}$, using DJL and kernel-based methods, for target policy $\pi(x)=x$ in the toy example.}\label{table:toy}
	\scalebox{0.88}{\begin{tabular}{lllll}
		\toprule
		Methods & Indicator&Deep Jump Learning&Kernel with $h=0.4$ & Kernel with $h=1$\\
		\midrule
		$ V^{(1)}(\pi)$&$\mathbb{I}\{\pi(X)\leq 0.5\}$ &0.31 (0.06) & 0.50 (0.08)& 0.40 (0.05) \\
		\midrule			
		$ V^{(2)}(\pi)$ &$\mathbb{I}\{\pi(X)> 0.5\}$& 0.09 (0.19) &0.16 (0.20) & 1.09 (0.09)  \\ 
		\bottomrule
	\end{tabular}}
\end{table}

To overcome this limitation, we propose to adaptively discretize the treatment space into a union of disjoint intervals such that within each interval {$\mathcal{I}$, the conditional mean function $Q$ can be well-approximated by some functions $q_{\mathcal{I}}$ that depend on features but not on the treatment (constant in $a$), i.e., $Q(\bullet,a)\approx\sum_{\mathcal{I} \in \mathcal{D}} \{\mathbb{I}(a\in \mathcal{I}) q_{\mathcal{I}}(\bullet)\}$}. 
By the discretization, one can apply the IPW or DR methods to evaluate the value. The advantage of adaptive discretization is illustrated in the right panel of Figure \ref{fig:toy}, where we apply the proposed DJL method to the toy example. See details of the proposed method and its implementation in Sections \ref{sec:mainidea} and \ref{sec:algo}. When $a\le0.5$, $Q(x,a)$ is constant in $a$. It is likely that our procedure will not further split the interval $[0,0.5]$. Consequently, the corresponding DR estimator for $V^{(1)}(\pi)$ will not suffer from large variance. When $a> 0.5$, our procedure will split $(0.5,1]$ into a series of sub-intervals, approximating $Q$ by a step function. This guarantees the resulting DR estimator for $V^{(2)}(\pi)$ will not suffer from large bias. Consequently, the proposed value estimator achieves a smaller mean squared error than kernel-based estimators. See Table \ref{table:toy} for details.

 \subsection{The Main Idea}\label{sec:mainidea}

We consider the following two model assumptions, which cover a variety of scenarios in practice. 



\textbf{Model 1: Piecewise function.} Suppose
	\begin{eqnarray}\label{step}
Q(x,a)=\sum_{\mathcal{I} \in \mathcal{D}_0} \left\{ q_{\mathcal{I},0}(x)\mathbb{I}(a \in \mathcal{I})\right\} , \quad  \text{ for any } x \in \mathcal{X},  \text{ for any } a \in  \mathcal{A},
	\end{eqnarray} 
for some partition $\mathcal{D}_0$ of $[0,1]$ and a collection of functions $\{q_{\mathcal{I},0}\}_{\mathcal{I}\in \mathcal{D}_0}$. 


\textbf{Model 2: Continuous function.} 
Suppose $Q$ is a continuous function of $a$ and $x$.  

{Model 1 covers the dynamic pricing example we mentioned in the introduction. In our simulation studies in Section \ref{sec:simu_part1}, the underlying model is set to be a piecewise function in Scenarios 1 and 2. Model 2 covers the personalized dose-finding example, Scenarios 3 and 4 in our simulation studies, as well as the real data section in Section \ref{sec:real}. } We next detail the proposed method, which will work when either Model 1 or 2 holds.

Motivated by Model 1, our goal is to identify an optimal discretization $\widehat{\mathcal{D}}$ such that for each interval $\mathcal{I}\in \widehat{\mathcal{D}}$, $Q(x,a)$ is approximately a constant function of $a\in \mathcal{I}$. {Specifically, under Model 1, we assume the function $Q(x,a)$ is a piecewise function on the action space. Within each segment $\mathcal{I}$, the function $Q(x,a)$ is a constant function of $a$, but can be any function of the features $x$. In other words, $Q(x,a_1) = Q(x,a_2)$ for any $a_1,a_2 \in \mathcal{I}$. Thus, we denote the function $Q(x,a)$ at each segment $\mathcal{I}$ as $q_\mathcal{I}(x)$, which yields a piecewise function $Q(x,a) = \sum_{\mathcal{I}} q_\mathcal{I}(x) \mathbb{I}(a \in \mathcal{I})$, as stated in \eqref{step}. In the real applications, the true function $Q(x,a)$ could be either a continuous function, or a piecewise function.  As such, we propose to approximate the underlying unknown function $Q(x,a)$ by these piecewise functions of $a$ using the proposed DJL method. 
Such approximation} allows us to derive the DR estimator based on $\widehat{\mathcal{D}}$. The bias and variance of the resulting estimator are largely affected by the number of intervals in $\widehat{\mathcal{D}}$. 
Specifically, if $|\widehat{\mathcal{D}}|$ is too small, then the piecewise approximation is not accurate, leading to a biased estimator. If  $|\widehat{\mathcal{D}}|$ is too large, 
then $\widehat{\mathcal{D}}$ will contain many short intervals, and the resulting estimator might suffer from a large variance. 

To this end, we develop a data-adaptive method to compute $\widehat{\mathcal{D}}$.
 We first divide the treatment space $\mathcal{A}$ into $m$ disjoint intervals: 
$
[0,1/m),[1/m,2/m)$, $\dots$, $[(m-1)/m,1] 
$.  
We require the integer $m$ to diverge with the sample size $n$,
such that the conditional mean function $Q$ can be well-approximated by a piecewise function on these intervals. Note that these $m$ initial intervals is not equal to $\widehat{\mathcal{D}}$, but only serve as the initial candidate intervals. Yet, $\widehat{\mathcal{D}}$ will be constructed by adaptively combining some of these intervals. 
We find in our numerical studies that the size of the final partition $|\widehat{\mathcal{D}}|$ is usually much less than $m$ (see Table \ref{table:add1} in Appendix \ref{add:exp_res} for more details).
In practice, we recommend to set the initial number of intervals $m$ to be proportional to the sample size $n$, i.e., $m=n/c$ for some constant $c>0$. The performance of the resulting value estimator is not overly sensitive to the choice of $c$. 

We define $\mathcal{B}(m)$ as the set of discretizations $\mathcal{D}$ such that 
each interval $\mathcal{I}\in \mathcal{D}$ corresponds to a union of some of the $m$ initial intervals. 
Each discretization $\mathcal{D}\in \mathcal{B}(m)$ is associated with a set of functions $\{q_{\mathcal{I}}\}_{\mathcal{I}\in \mathcal{D}}$. {
We model these $q_{\mathcal{I}}$ using DNNs, to capture the complex dependence between the outcome and features. 
When $Q(\bullet,a)$ is well-approximated by $\sum_{\mathcal{I} \in \mathcal{D}} \{\mathbb{I}(a\in \mathcal{I}) q_{\mathcal{I}}(\bullet)\}$}, we expect the least square loss
$
\sum_{\mathcal{I}\in \mathcal{D}}    \sum_{i=1}^{n}  [\mathbb{I}(A_i\in \mathcal{I})  \big\{Y_i - q_{\mathcal{I}}(X_i) \big\}^2 ] ,
$
will be small.
Thus, $\widehat{\mathcal{D}}$ can be estimated by solving 
{\begin{eqnarray}\label{eqn:optimize}
 \left(\widehat{\mathcal{D}},\{\widehat{q}_{\mathcal{I}}:\mathcal{I}\in \widehat{\mathcal{D}} \}\right)=\argmin_{\left(\substack{\mathcal{D}\in \mathcal{B}(m),\\ \{q_{\mathcal{I}}\in \mathcal{Q}_{\mathcal{I}}: \mathcal{I}\in \mathcal{D} \} }\right)}\left(\sum_{\mathcal{I}\in \mathcal{D}} \left[ {1\over n} \sum_{i=1}^{n} \mathbb{I}(A_i\in \mathcal{I})  \big\{Y_i - q_{\mathcal{I}}(X_i) \big\}^2\right]+\gamma_n |\mathcal{D}| \right),
\end{eqnarray}}
for some regularization parameter $\gamma_n$ and some function class of DNNs $\mathcal{Q}_{\mathcal{I}}$.
Here, the penalty term $\gamma_n |\mathcal{D}|$ in 
\eqref{eqn:optimize} controls the total number of intervals in $\widehat{\mathcal{D}}$, as in multi-scale change point detection. A large $\gamma_n$ results in few intervals in $\widehat{\mathcal{D}}$ and a potential large bias of the value estimator, whereas a small $\gamma_n$ procedures a large number of intervals in $\widehat{\mathcal{D}}$, leading to a noisy value estimator. The theoretical order of $\gamma_n$ is detailed in Section \ref{rate}. In practice, we use cross-validation to select $\gamma_n$. We refer to this step as deep discretization. Details of solving \eqref{eqn:optimize} are given in Section \ref{sec:algo}.

Given $\widehat{\mathcal{D}}$ and $\{\widehat{q}_{\mathcal{I}}:\mathcal{I}\in \widehat{\mathcal{D}} \}$, we apply the DR estimator in \eqref{eqn:DR} to derive the value estimate for any target policy of interest $\pi$, i.e., 
\begin{eqnarray}\label{value_djqe}
	\widehat{V}^{DR}(\pi)={1\over n} \sum_{\mathcal{I}\in \widehat{\mathcal{D}}}  \sum_{i=1}^{n} \left( \mathbb{I}\{\pi(X_i)\in \mathcal{I}\}  \left[  {\mathbb{I}(A_i\in \mathcal{I})\over  \widehat{b}_\mathcal{I}(X_i)}\big\{Y_i - \widehat{q}_\mathcal{I}(X_i) \big\} + \widehat{q}_\mathcal{I}(X_i) \right]\right),
\end{eqnarray} 
where $\widehat{b}_{\mathcal{I}}(x)$ is some estimator of the generalized propensity score function $\prob(A\in \mathcal{I}|X=x)$. 
We call this method as the deep jump learning. 
{We remark that the proposed method yields a consistent value estimator allowing the function $Q$ to be either a continuous or piecewise function of the treatment. Under different model assumptions, we derive the corresponding rate of convergence of our method in Section \ref{rate}.}

\subsection{The Complete Algorithm for Deep Jump Learning}\label{sec:algo}
We present the details for DJL in this section. To further reduce the bias of the value estimator in \eqref{value_djqe}, we employ a data splitting and cross-fitting strategy \citep{chernozhukov2017double}. That is, we use different subsets of data samples to estimate the discretization $\widehat{\mathcal{D}}$ and to construct the value estimator. 
Our algorithm consists of three steps: data splitting, deep discretization, and cross-fitting. We detail each of these steps below. 


%

\textbf{Step 1: data splitting.} We divide all $n$ samples into $\mathcal{L}$ {disjoint} subsets of equal size, where $\mathbb{L}_{\ell}$ denotes the indices of samples in the $\ell$th subset for $\ell=1,\cdots,\mathcal{L}$. Let $\mathbb{L}_{\ell}^c=\{1,2,\cdots,n\}-\mathbb{L}_{\ell}$ as the complement of $\mathbb{L}_{\ell}$. {Data splitting allows us to use one part of the data, i.e., $\mathbb{L}_{\ell}^c$, to train machine learning models for the conditional mean function and propensity score function, and the remaining part, i.e., $\mathbb{L}_{\ell}$, to estimate the value.
We aggregate the resulting estimates over different subsets to get full efficiency, as summarized in the third step}. 

\begin{algorithm}[!t] 
	\caption{{Deep Jump Learning}}\label{alg1}
	\begin{tabbing}
		\enspace \textbf{Global:} data $\{(X_i, A_i,Y_i )\}_{1\leq i\leq n}$; number of initial intervals $m$; penalty term $\gamma_n$;  \\\qquad \quad ~ target policy $\pi$.\\
		\enspace \textbf{Local:} 
		Bellman function ${\text{Bell}} \in  \mathbb{R}^m$; partitions $\widehat{\mathcal{D}}$;  DNN functions $\{\widehat{q}_{\mathcal{I}}, \widehat{b}_{\mathcal{I}}:\mathcal{I}\in \widehat{\mathcal{D}} \}$;  \\\qquad \quad ~  a vector $\tau\in \mathbb{N}^{m}$;   a set of candidate point lists $\mathcal{R}$.\\
		\enspace \textbf{Output:} 
		the value estimator for target policy $\widehat{V}(\pi)$.\\
		
		\enspace I. Split all $n$ samples into $\mathcal{L}$ subsets as $\{\mathbb{L}_{1},\cdots,\mathbb{L}_\mathcal{L}\}$; $\widehat{V}(\pi)\gets 0$;  \\		
		\enspace II.  Initialization:\\
		\qquad 1. Set even segment on the action space with $m$ pieces:\\
		\qquad  \quad $\{\mathcal{I}\}=\{[0,1/m) , [1/m,2/m) ,  \dots, [(m-1)/m,1]\}$;\\
		\qquad 2. Create a function to calculate cost $\mathcal{C}$ with inputs $(l,r)$: \\
		\qquad  \quad If $\mathcal{C}(l,r) ==NULL$:\\
		\qquad \quad (i). Let $\mathcal{I}=[l/m,r/m)$ if $r<m$ else $\mathcal{I}=[l/m,1]$;\\
		\qquad \quad (ii). Fit a DNN regressor: $\widehat{q}_{\mathcal{I} }(\cdot) \gets \mathbb{I}(i \in\mathbb{L}_{\ell}^c) \mathbb{I}(A_i\in \mathcal{I}  ) Y_i \sim \mathbb{I}(A_i\in \mathcal{I} )DNN(X_i) $;\\
		\qquad \quad (iii). Store the cost: $\mathcal{C}(\mathcal{I})  \gets  \sum_{i \in \mathbb{L}_{\ell}^c}\mathbb{I}(A_i\in\mathcal{I} ) \big\{\widehat{q}_{\mathcal{I} }(X_i )-Y_i\big\}^2$;\\
		\qquad \quad Return $\mathcal{C}(l,r)$;\\
		\enspace III. For $\ell=1,\cdots,\mathcal{L}$:\\
		\qquad 1. Set the training dataset as $ \mathbb{L}_{\ell}^c=\{1,2,\cdots,n\}-\mathbb{L}_{\ell}$;\\
		\qquad 2. $\text{Bell}(0)\gets -\gamma_n$; $\widehat{\mathcal{D}} =[0,1]$; $\tau \gets Null$; $\mathcal{R}(0) \gets \{0\}$;\\
		\qquad 3.  Apply the pruned exact linear time method to get partitions: For $v^*=1,\dots,m$:\\
		\qquad \quad (i).$\text{Bell}(v^*)=\min_{v\in \mathcal{R}{(v^*)}} \{\text{Bell}(v)+\mathcal{C}([v/m,v^*/m))+ \gamma_n\}$;\\
		\qquad \quad (ii). $v^1\gets \argmin_{v\in \mathcal{R}(v^*)} \{\text{Bell}(v)+\mathcal{C}([v/m,v^*/m))+ \gamma_n\}$;\\
		\qquad \quad (iii). $\tau(v^*)\gets \{v^1,\tau(v^1)\}$;\\ 
		\qquad \quad (iv). $\mathcal{R}(v^*) \gets \{v\in \mathcal{R}(v^*-1) \cup\{v^*-1\}: \text{Bell}(v)+\mathcal{C}([v/m,(v^*-1)/m))\leq \text{Bell}(v^*-1)\} $;\\
		\qquad 4. Construct the DR value estimator: $r\gets m$; $l\gets \tau[r] $; While $r>0$:\\
		\qquad \quad (i) Let $\mathcal{I}=[l/m,r/m)$ if $r<m$ else $\mathcal{I}=[l/m,1]$; $\widehat{\mathcal{D}}\gets \widehat{\mathcal{D}}\cup \mathcal{I} $;\\
		\qquad \quad (ii)  Recall fitted DNN: $\widehat{q}_{\mathcal{I} }(\cdot) \gets  \mathbb{I}(i \in\mathbb{L}_{\ell}^c) \mathbb{I}(A_i\in \mathcal{I} ) Y_i \sim \mathbb{I}(A_i\in \mathcal{I})DNN(X_i) $;\\
		\qquad \quad (iii) Fit propensity score: $\widehat{b}_{\mathcal{I} }(\cdot) \gets \mathbb{I}(i \in\mathbb{L}_{\ell}^c) \mathbb{I}(A_i\in \mathcal{I} ) \sim \mathbb{I}(A_i\in \mathcal{I} )DNN(X_i) $;\\
		\qquad \quad (iv)  $r\gets l$; $l\gets \tau(r) $;\\
		\qquad 6. Evaluation using testing dataset  $ \mathbb{L}_{\ell}$:\\
		\qquad  \quad $\widehat{V}(\pi) += \sum_{\mathcal{I}\in \widehat{\mathcal{D}}} \left(   \sum_{i \in \mathbb{L}_{\ell} } \mathbb{I}(A_i\in \mathcal{I})  \left[  {\mathbb{I}\{\pi (X_i)\in \mathcal{I}\} \over  \widehat{b}_\mathcal{I}( X_i)}\big\{Y_i - \widehat{q}_\mathcal{I}(X_i ) \big\} + \widehat{q}_\mathcal{I}(X_i ) \right]\right)$;\\
		
		\enspace \Return $\widehat{V}(\pi)/n$ .
	\end{tabbing}
\end{algorithm}

\textbf{Step 2: deep discretization.} For each $\ell=1,\cdots,\mathcal{L}$, we propose to apply deep discretization to compute a discretization $\widehat{\mathcal{D}}^{(\ell)}$ and $\{\widehat{q}_{\mathcal{I}}^{(\ell)}:\mathcal{I}\in\widehat{\mathcal{D}}^{(\ell)} \}$ by solving a version of \eqref{eqn:optimize} using the data subset in $\mathbb{L}_{\ell}^c$ only. We next present the computational details for solving this optimization. Our algorithm {employs the pruned exact linear time method \citep{killick2012optimal} to identify the change points with a cost function that involves DNN training.} 
Specifically, for any interval $\mathcal{I}$, define $\widehat{q}_{\mathcal{I}}^{(\ell)}$ as the minimizer of
\begin{eqnarray}\label{eqn:cost_cf}
 \argmin_{q_{\mathcal{I}}\in \mathcal{Q}_{\mathcal{I}}}\frac{1}{|\mathbb{L}_{\ell}^c|}\sum_{i \in \mathbb{L}_{\ell}^c} \left[\mathbb{I}(A_i\in\mathcal{I} ) \big\{q_{\mathcal{I}}(X_i )-Y_i\big\}^2\right],
\end{eqnarray}
where $|\mathbb{L}_{\ell}^c|$ denotes the number of samples in $\mathbb{L}_{\ell}^c$. Define the cost function $\mathcal{C}^{(\ell)}(\mathcal{I})$ as the minimum value of the objective function \eqref{eqn:cost_cf}, i.e, 
\begin{eqnarray*}
\mathcal{C}^{(\ell)}(\mathcal{I}) = \frac{1}{|\mathbb{L}_{\ell}^c|}\sum_{i \in \mathbb{L}_{\ell}^c}\left[\mathbb{I}(A_i\in\mathcal{I} ) \big\{\widehat{q}_{\mathcal{I}}^{(\ell)}(X_i )-Y_i\big\}^2\right].
\end{eqnarray*}
Computation of $\widehat{\mathcal{D}}^{(\ell)}$ relies on dynamic programming \citep{friedrich2008complexity}. For any integer $1\le v^*<m$, denote by $\mathcal{B}(m, v^*)$ the set consisting of all possible discretizations $\mathcal{D}_{v^*}$ of $[0,v^*/m)$. Set $\mathcal{B}(m,m)=\mathcal{B}(m)$, we define the Bellman function as 
\begin{eqnarray*}
	\textrm{Bell}(v^*)=\inf_{\mathcal{D}_{v^*}\in \mathcal{B}(m,v^*)}\left\{\sum_{\mathcal{I}\in \mathcal{D}_{v^*}} \mathcal{C}^{(\ell)}(\mathcal{I})+\gamma_n (|\mathcal{D}_{v^*}|-1) \right\}, \text{ and  } \textrm{Bell}(0)=-\gamma_n.
\end{eqnarray*}
{Our algorithm recursively updates the Bellman cost function for $v^*=1,2,\cdots$ by
\begin{eqnarray}\label{bellmanrelation}
	\textrm{Bell}(v^*)=\min_{v\in \mathcal{R}_{v^*}} \left\{\textrm{Bell}(v)+\mathcal{C}^{(\ell)}([v/m,v^*/m)) +\gamma_n \right\},\,\,\,\,\,\,\,\,\text{ for any } v^*\ge 1,
\end{eqnarray}
where $\mathcal{R}_{v^*}$ is the candidate change points list. For a given $v$, the right-hand-side of \eqref{bellmanrelation} corresponds to the cost of partitioning on a particular point. We then identify the best $v$ that minimizes the cost. This yields the Bellman function on $[0, v^*/m]$ on the left-hand-side. In other words, \eqref{bellmanrelation} is a recursive formula used in our dynamic algorithm to update the Bellman equation for the locations of change points. It is recursive as the Bellman function appears on both sides of \eqref{bellmanrelation}. Here, the list of candidate change points $\mathcal{R}_{v^*}$ is given by 
\begin{eqnarray}\label{Rvstar}
\left\{v\in \mathcal{R}_{v^*-1} \cup\{v^*-1\}: \textrm{Bell}(v)+\mathcal{C}^{(\ell)}([v/m,(v^*-1)/m))\leq \textrm{Bell}(v^*-1)\right\},
\end{eqnarray}
during each iteration with $\mathcal{R}_0=\{0\}$. The constraint listed in \eqref{Rvstar} restricts the research space in \eqref{bellmanrelation} to a potentially much smaller set of candidate change points, i.e., $\mathcal{R}_{v*}$. The main purpose is to facilitate the computation by discarding change points not relevant to obtain the final discretization. It yields a linear computational cost \citep{killick2012optimal}. In contrast, without these restrictions, it would yield a quadratic computational cost \citep{friedrich2008complexity}.}

To solve \eqref{bellmanrelation}, we search the optimal change point location $v$ that minimizes $\textrm{Bell}(v^*)$. This requires deep learning to estimate $\widehat{q}_{\mathcal{I}}^{(\ell)}$ and $\mathcal{C}^{(\ell)}(\mathcal{I} )$ with $\mathcal{I} = [v/m,v^*/m)$ for each $v\in \mathcal{R}_{v^*}$. 
Let $v^1$ be the corresponding minimizer. {
We then define the change points list $\tau(v^*)$ as the set of change-point locations in $[0, v^*/m]$ computed by the dynamic programming algorithm. It is computed iteratively based on the update $\tau(v^*)=\{v^1, \tau(v^1)\}$, which means that during each iteration, it includes the current best change point location $v^1$ (that minimizes \eqref{bellmanrelation}) and the previous change-point list for the interval $[0,v^1/m]$. 
This procedure is iterated to compute $\textrm{Bell}(v^*)$ and $\tau(v^*)$ for $v^*=1,\dots,m$, to find the best change-point set for interval $[0,1]$.} The optimal partition $\widehat{\mathcal{D}}^{(\ell)}$ is determined by the values stored in $\tau$. Specifically, we initialize $\widehat{\mathcal{D}}^{(\ell)}=[\tau(m)/m,1]$, $r=m$ and recursively update $\widehat{\mathcal{D}}^{(\ell)}$ by setting $\widehat{\mathcal{D}}^{(\ell)}\leftarrow \widehat{\mathcal{D}}^{(\ell)}\cup [\tau(r)/m,r/m)$ and $r\leftarrow \tau(r)$, as in dynamic programming \citep{friedrich2008complexity}.

\textbf{Step 3: cross-fitting.} For each interval in the estimated optimal partition $\widehat{\mathcal{D}}^{(\ell)}$, let $\widehat{b}_{\mathcal{I}}^{(\ell)}(x)$ denote some estimator for the propensity score $\prob(A\in \mathcal{I}|X=x)$. In a randomized study, the density function $b(a|x)$ is known to us and we set $\widehat{b}_{\mathcal{I}}^{(\ell)}(x)=\int_{a\in \mathcal{I}}b(a|x)da$. To deal with data from observational studies, we estimate the generalized propensity score with deep learning using the training dataset $\mathbb{L}_{\ell}^c$ as $\widehat{b}^{(\ell)}_{\mathcal{I}}(x)$. 
{We evaluate the target policy in each subsample $\mathbb{L}_\ell$, based on the estimators ($\widehat{q}_\mathcal{I}^{(\ell)}$, $\widehat{b}_\mathcal{I}^{(\ell)}$, and $\widehat{\mathcal{D}}^{(\ell)}$) trained in its complementary subsamples $\mathbb{L}_{\ell}^c = \{1,\cdots,n\} - \mathbb{L}_\ell $. Denote this value estimator for subset $\mathbb{L}_\ell$ as $\widehat{V}_\ell$. The final proposed value estimator for $V(\pi)$ is to aggregate over $\widehat{V}_\ell$ for $\ell =1 ,\cdots,\mathcal{L}$ via cross-fitting, } given by,
\begin{eqnarray}\label{value_cf}
	\widehat{V}(\pi)={1\over n} \sum_{\ell=1}^\mathcal{L} \sum_{\mathcal{I}\in \widehat{\mathcal{D}}^{(\ell)}}    \sum_{i \in  \mathbb{L}_{\ell}}   \left[ \mathbb{I}(A_i\in \mathcal{I}) {\mathbb{I}\{\pi(X_i)\in \mathcal{I}\} \over  \widehat{b}^{(\ell)}_\mathcal{I}(X_i)}\big\{Y_i - \widehat{q}_{\mathcal{I}}^{(\ell)}(X_i) \big\} +\mathbb{I}(A_i\in \mathcal{I}) \widehat{q}_{\mathcal{I}}^{(\ell)}(X_i) \right]. 
\end{eqnarray}
Note the samples used to construct $\widehat{V}$ inside bracket are independent from those to estimate $\widehat{q}_{\mathcal{I}}^{(\ell)}$, $\widehat{b}_{\mathcal{I}}^{(\ell)}$ and $\widehat{\mathcal{D}}^{(\ell)}$. This helps remove the bias induced by overfitting in the estimation of $\widehat{q}_{\mathcal{I}}^{(\ell)}$, $\widehat{b}_{\mathcal{I}}^{(\ell)}$ and $\widehat{\mathcal{D}}^{(\ell)}$.

{We give the full detailed pseudocode in Algorithm \ref{alg1}. The computational complexity required to implement the proposed approach is $\mathcal{O}(mC_n)$, where $C_n$ is the computational complexity of training one DNN model with the sample size $n$. Detailed analysis is provided in Section \ref{add:compu_complex} in Appendix. The code is publicly available at our repository at \url{https://github.com/HengruiCai/DJL}.} 

\section{Theory}\label{rate}


We investigate the theoretical properties of the proposed DJL method. All the proofs are provided in the supplementary article. Without loss of generality, assume the support $\mathcal{X}=[0,1]^p$. To simplify the analysis, we focus on the case where the behavior policy $b$ is known to us, 
which automatically holds for data from randomized studies. 
We focus on the setting where the conditional mean function $Q$ is a smooth function of the features; see A1 below. Specifically, define the class of $\beta$-smooth functions, also known as H{\"o}lder smooth functions with exponent $\beta$, as 
\begin{eqnarray*}
	\Phi(\beta,c)=\left\{h:\sup_{\|\alpha\|_1\le \floor{\beta}} \sup_{x\in \mathcal{X}} |\Delta^{\alpha} h(x)|\le c, \sup_{\|\alpha\|_1=\floor{\beta}} \sup_{\substack{x,z\in \mathcal{X}\\ x\neq z}} \frac{|\Delta^{\alpha} h(x)-\Delta^{\alpha} h(z)|}{\|x-z\|_2^{\beta-\floor{\beta}}}\le c \right\},
\end{eqnarray*} 
for some constant $c>0$, where $\floor{\beta}$ denotes the largest integer that is smaller than $\beta$ and $\Delta^{\alpha}$ denotes the differential operator 
$\Delta^{\alpha}$ denote the differential operator:
$
	\Delta^{\alpha}h(x)= {\partial^{\|\alpha\|_1} h(x)}/{\partial x_1^{\alpha_1}\cdots\partial x_p^{\alpha_p}},
$ 
where $x=[x_1,\dots,x_p]$. 
When $\beta$ is an integer, $\beta$-smoothness essentially requires a function to have bounded derivatives up to the $\beta$th order. {The H{\"o}lder smoothness assumption is commonly imposed in the current literature \citep[see e.g., ][]{farrell2021deep}, which is a special example of the function classes that can be learned by neural nets. Meanwhile, the proposed DJL method is valid when $Q(x,a)$ is a nonsmooth function of $x$ as well \citep[see e.g.,][]{imaizumi2019deep}. Our theory thus can be further generalized to any function class that can be learned by neural nets at a certain rate.} 
We introduce the following conditions.


\noindent (A1.) Suppose $b(a|\bullet)\in \Phi(\beta,c)$, and $Q(\bullet,a)\in \Phi(\beta,c)$ for any $a$. \\
\noindent (A2.) Functions $\{\widehat{q}_{\mathcal{I}}:{\mathcal{I}\in \widehat{\mathcal{D}}^{(\ell)} \} }$ are uniformly bounded.

Assumption (A2) ensures that the optimizer would not diverge in the uniform norm sense. Similar assumptions are commonly imposed in the literature to derive the convergence rates of DNN estimators \citep[see e.g.,][]{farrell2021deep}. Combining (A2) with (A1) allows us to derive the uniform rate of convergence for the class of DNN estimators $\{\widehat{q}_{\mathcal{I}}: {\mathcal{I}\in \widehat{\mathcal{P}}}\}$. Specifically, $\widehat{q}_{\mathcal{I}}$ converges at a rate of $O_p\{n|\mathcal{I}|^{-2\beta/(2\beta+p)}\}$ where the big-$O$ terms are uniform in $\mathcal{I}$, $p$ is the dimension of features. See Lemma \ref{lemma0} in the supplementary article for details.
\subsection{Properties under Model 1}
We first consider Model 1 where the function $Q(x,a)$ takes the form of \eqref{step}. As commented, this assumption holds in applications such as dynamic pricing. 
Without loss of generality, assume $q_{\mathcal{I}_1,0}\neq q_{\mathcal{I}_2,0}$ for any two adjacent intervals $\mathcal{I}_1,\mathcal{I}_2\in \mathcal{D}_0$. This guarantees that the representation in \eqref{step} is unique. Let $L_{\mathcal{I}}$ and $W_{\mathcal{I}}$ be the number of hidden layers and total number of parameters in the function class of DNNs $\mathcal{Q}_{\mathcal{I}}$. Assume the number of change points in $\mathcal{D}_0$ is fixed. 
The following theorem summarizes the rate of convergence of the proposed estimator under Model 1.
\begin{theorem}\label{thm1}
	Suppose \eqref{step}, (A1) and (A2) hold. Suppose $m$ is proportional to $n$, $Y$ is a bounded variable and 
	$A$ has a bounded probability density function on $[0,1]$. Assume $\{\gamma_n\}_{n \in \mathbb{N}}$ satisfies $\gamma_n \to 0$ and $\gamma_n \gg n^{-2\beta/(2\beta+p)}\log^8n$. Then, there exist some classes of DNNs $\{\mathcal{Q}_{\mathcal{I}}:\mathcal{I}\}$ with $L_{\mathcal{I}}\asymp \log (n|\mathcal{I}|)$ and $W_{\mathcal{I}}\asymp (n|\mathcal{I}|)^{p/(2\beta+p)}\log (n|\mathcal{I}|)$ such that 
	the following events occur with probability at least $1-O(n^{-2})$,\\ 
	\noindent (i) $|\widehat{\mathcal{D}}^{(\ell)}|=|\mathcal{D}_0|$;  
	and (ii) $\max_{\tau\in J(\mathcal{D}_0)} \min_{\widehat{\tau}\in J(\widehat{\mathcal{D}}^{(\ell)})} |\widehat{\tau}-\tau|=O\{n^{-2\beta/(2\beta+p)}\log^8 n\}$. \\
	In addition, for any policy $\pi$ such that for any $\tau_0\in J(\mathcal{D}_0)$, $\prob\{\pi(X)\in [\tau_0-\epsilon,\tau_0+\epsilon]\}=O(\epsilon)$, \\
	(iii) $\widehat{V}(\pi)=V(\pi)+O_p\{n^{-2\beta/(2\beta+p)}\log^8 n\}+O_p(n^{-1/2})$. 

\end{theorem} 

We make a few remarks. First, the result in (i) imply that deep discretization correctly identifies the number of change points. The result in (ii) imply that any change point in $\mathcal{D}_0$ can be consistently identified. 
In particular, $J(\widehat{\mathcal{D}}^{(\ell)})$ corresponds to a subset of 
$\{1/m,2/m,\cdots,(m-1)/m\}$. 
For any true change point $\tau$ in $\mathcal{D}_0$, there will be a change point in $\widehat{\mathcal{D}}^{(\ell)}$ that approaches $\tau$ at a rate of $n^{-2\beta/(2\beta+p)}$ up to some logarithmic factors.  Second, it can be seen from the proof of Theorem \ref{thm1} that the two error terms $O\{n^{-2\beta/(2\beta+p)}\log^8 n\}$ and $O(n^{-1/2})$ in (iii) correspond to the bias and standard deviation of the proposed value estimator, respectively. When $2\beta>p$, the bias term is negligible. A Wald-type confidence interval can be constructed to infer $V(\pi)$. The assumption $2\beta>p$ allow the deep learning estimator to converge at a rate faster than $n^{-1/4}$. Such a condition is commonly imposed in the literature for inferring the average treatment effect \citep[see e.g.,][]{chernozhukov2017double,farrell2021deep}.  {When $\beta<p$, i.e., the underlying conditional mean function $Q$ is not smooth enough, the proposed estimator suffers from a large bias and might converge at a rate that is slower than the usual parametric rate. This concern can be addressed by employing the A-learning method \citep[see e.g., ][]{Murphy:2003aa,schulte2014q,shi2018high}. The A-learning method is more robust and requires weaker conditions to achieve the parametric rate. Specifically, it only requires the difference $Q(x,1)-Q(x,0)$ to belongs to $\Phi(\beta,c)$. This is weaker than requiring both $Q(x,1)$ and $Q(x,0)$ to belongs to $\Phi(\beta,c)$.}  Third, to ensure the consistency of the proposed value estimator, we require that the distribution of the random variable $\pi(X)$ induced by the target policy does not have point-masses at the change point locations. This condition is mild. For nondynamic policies where $\pi(X)=\pi_0$ almost surely, it requires $\pi_0\notin J(\mathcal{D}_0)$. We remark that the set $J(\mathcal{D}_0)$ has a zero Lebesgue measure on $[0,1]$. For dynamic policies, it automatically holds when $\pi(X)$ has a bounded density on $[0,1]$.

\subsection{Properties under Model 2}
We next consider Model 2 where the function $Q(x,a)$ is continuous in the treatment space. 
\begin{theorem}\label{thm2}
	Assume $Q(x,a)$ is Lipschitz continuous, i.e., $|Q(x,a_1)-Q(x,a_2)|\le L|a_1-a_2|$, for all $a_1,a_2\in [0,1], x\in \mathcal{X}$, and some constant $L>0$.  Assume (A1) and (A2), and $m$ is proportional to $n$ and $\gamma_n$ is proportional to $\max\{n^{-3/5},n^{-2\beta/(2\beta+p)}\log^9 n\}$. Then for any target policy $\pi$, 
	\begin{eqnarray*}\widehat{V}(\pi)-V(\pi)=O_p(n^{-1/5})+O_p\{n^{-2\beta/(6\beta+3p)}\log^3 n\}.\end{eqnarray*}\end{theorem}
When $4\beta>3p$, the convergence rate is given by $O_p(n^{-1/5})$. We remark that the above upper bound is valid for any target policy $\pi$.
{The convergence rate in Theorem \ref{thm2} may not be tight. To the best of our knowledge, no formal lower bounds of the value estimator have been established in the literature in the continuous treatment setting. In the literature on multi-scale change point detection, there are lower bounds on the estimated time series \citep[see e.g., ][]{boysen2009consistencies}. However, they considered settings without baseline covariates and it remains unclear how the rate of convergence of the estimated piecewise function can be translated into that of the value. We leave this for future research.}

 {Finally, we clarify our theoretical contributions compared with the deep learning theory established in \citet{farrell2021deep}. First, \citet{farrell2021deep} considered a single DNN, whereas we established the uniform convergence rate of several DNN estimators, since our proposal requires to train multiple DNN models. Establishing the uniform rate of convergence poses some unique challenges in deriving the results of Theorems \ref{thm1} and \ref{thm2}. We need to control the initial number of the intervals $m$ to be proportional to $n$ and the order of penalty term $\gamma_n$, so that uniform convergence rate can be established across all intervals. To address this difficulty, we derive the tail inequality to bound the rate of convergence of the DNN estimator and use the Bonferroni's correction to establish the uniform rate of convergence.}
\subsection{Comparison with Kernel-Based Methods}\label{sec:kernel}
To simplify the analysis, we assume the kernel function is symmetric, the nuisance function estimators $\widehat{Q}$ and $\widehat{b}$ are set to their oracle values $Q$ and $b$, and that $4\beta>3p$. Suppose Model 1 holds. In Appendix \ref{eqn:ratekernel}, we show that the convergence rate of kernel-based methods is given by $O_p(n^{-1/3})$ with optimal bandwidth selection. In contrast, the proposed estimator converges at a faster rate of $O_p(n^{-1/2})$. Suppose Model 2 holds. In Appendix \ref{eqn:ratekernel}, we show that the convergence rate of kernel-based methods is given by $O_p(h)+O_p( {n^{-1/2}h^{-1/2}})$. 
Thus, kernel-based estimators converge at a slower rate when the bandwidth undersmoothes or oversmoothes the data.
In addition, as we have commented in Section \ref{sec:toy}, in cases where the second-order derivative of $Q$ has an abrupt change in the treatment space, kernel-based methods suffer from either a large bias, or a large variance.
Specifically, when $h$ is either much larger than $n^{-1/5}$ or much smaller than $n^{-3/5}$, our estimator converges at a faster rate. Kernel-based estimators could converge at a faster rate when $Q$ has a uniform degree of smoothness over the entire treatment space and the optimal bandwidth parameter is correctly identified.

\section{Simulation Studies}\label{sec:simu}
In this section, we investigate the finite sample performance of the proposed method on the simulated and real datasets, in comparison to three kernel-based methods. The computing infrastructure used is a virtual machine in the AWS Platform with 72 processor cores and 144GB memory. 


\subsection{Simulation Settings}\label{sec:simu_part1}
Simulated data are generated from the following model: \\
$
	Y|X,A \sim N\{Q(X,A), 1\},\,\,\,\, b(A|X) \sim \hbox{Uniform}[0,1]\,\,\,\,\hbox{and}\,\,\,\,X^{(1)},\dots,X^{(p)}\stackrel{iid}{\sim} \hbox{Uniform}[-1,1], $\\
where $X=[X^{(1)},\cdots,X^{(p)}]$. Consider the following different scenarios: \\
  \noindent \textbf{S1}:  $Q(x,a)=  (1+x^{(1)}) \mathbb{I}(a<0.35) + (x^{(1)}- x^{(2)})\mathbb{I}(0.35\le a<0.65) + (1- x^{(2)})\mathbb{I}(a\ge 0.65);$\\
\noindent \textbf{S2}:
	$Q(x,a)=\mathbb{I}(a <0.25) + \sin (2\pi x^{(1)})\mathbb{I}(0.25\le a<0.5) + \{0.5 - 8(x^{(1)}-0.75)^2\}\mathbb{I}(0.5\le a<0.75) +0.5 \mathbb{I}( a\ge0.75);$\\
\noindent \textbf{S3 (toy example)}:
	$Q(x,a) =  10 \max\{a^2-0.25,0\} \log(x^{(1)}+2) $;\\
\noindent \textbf{S4}:
	$Q(x,a) = 0.2(8 + 4x^{(1)} - 2x^{(2)} - 2x^{(3)}) - 2( 1 + 0.5x^{(1)} + 0.5x^{(2)} - 2a)^2.$\\
The function $Q(x,a)$ is a piecewise function of $a$ under Scenarios 1 and 2, and is continuous under Scenarios 3 (toy example considered in Section \ref{sec:toy}) and  4. We set the target policy to be the optimal policy that achieves the highest possible mean outcome. 
The dimension of the features is fixed to $p=20$. We consider four choices of the sample size, corresponding to $n=50, 100,200$ or $300$.

 \begin{figure}[!t]
	\centering
	\includegraphics[width=1\textwidth]{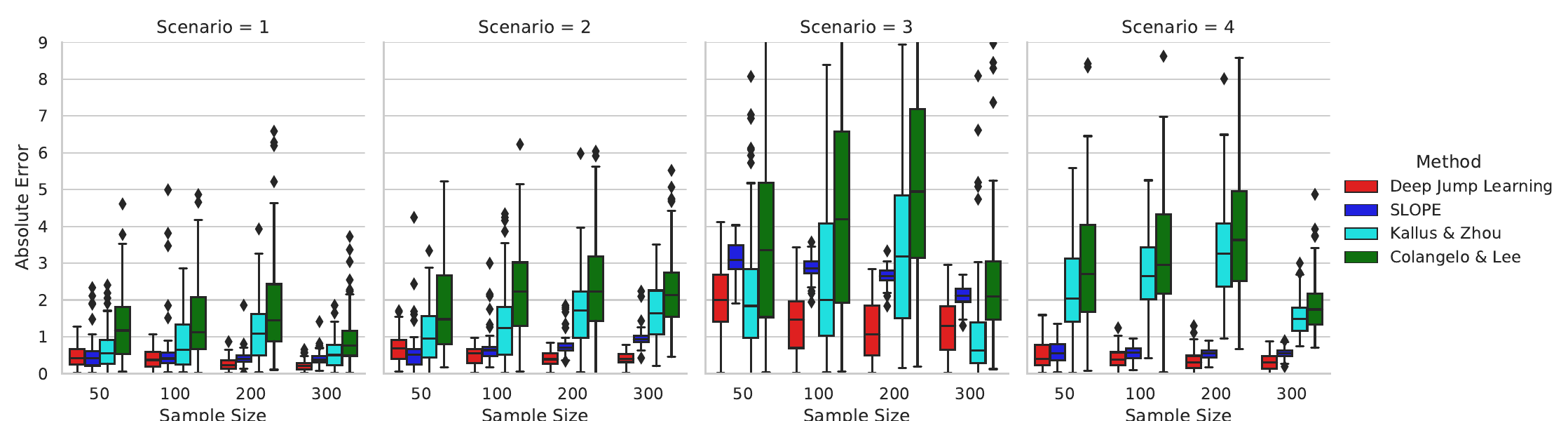}
	\caption{The box plot of the estimated values under the optimal policy via the proposed DJL method and three kernel-based methods for Scenario 1-4. The target values are 1.33, 1, 4.86 and 1.6, respectively.}\label{fig:simu}
\end{figure}

We compare the proposed DJL method with three kernel-based methods \citep{kallus2018policy,colangelo2020double,su2020adaptive}. In our implementation, we set $\mathcal{Q}_{\mathcal{I}}$ to the class of multilayer perceptrons (MLP) for each $\mathcal{I}$. This is a commonly used architecture in deep learning \citep{farrell2021deep}. 
The optimization in \eqref{eqn:cost_cf} is solved via the MLP regressor implemented by \citet{SCIKit2011} using a stochastic gradient descent algorithm, with tuning parameters set to the default values.  {In addition, we estimate the propensity score function using MLP as well.}
We set $m=n/10$ to achieve a good balance between the absolute error and the computational cost (see Figure \ref{fig:add1} in Appendix \ref{add:exp_res} for details).  
The averaged computational time are summarized in Table \ref{table:add2} {with additional results under large sample sizes $n=1000 \sim10000$ in Table \ref{table:add_rebuttal},  in Appendix \ref{add:exp_res}}. Overall, it takes a few minutes (less than 1 min for $n=50$ and 14 mins for $n=300$) to implement DJL, whereas the runtime of \citet{kallus2018policy}'s method is 365 mins for sample size $n=50$ and over 48 hours for $n=300$. Thus, as suggested in \citet{kallus2018policy}, to implement their method, we first compute $h^*$ using data with sample size $n_0=50$. To accommodate data with different $n$, we adjust $h^*$ by setting $h^*\{{n_0 /  n}\}^{0.2}$. To implement \citet{colangelo2020double}'s estimator, we consider a list of bandwidths suggested in their paper, given by $h = c\sigma_A n^{-0.2}$ with $c \in \{0.5, 0.75 , 1.0,  1.5\}$ and $\sigma_A$ is the standard deviation of the treatment. We then manually select the best bandwidth such that the resulting value estimator achieves the smallest mean squared error. The kernel-based method (SLOPE) by \citet{su2020adaptive} adopted the Lepski's method for bandwidth selection. In their implementation, they used the IPW estimator to evaluate the value. For a fair comparison, we replace it with DR to make the resulting estimator more efficient.


The average estimated value and its standard deviation over 100 replicates are illustrated in Figure \ref{fig:simu} for different methods, with detailed values reported in Table \ref{table:1} in Appendix \ref{add:exp_res}. In addition, we provide the size of the final estimated partition under DJL in Table \ref{table:add1} in Appendix \ref{add:exp_res}, which is much smaller than $m$ in most cases. It can be seen from Figure \ref{fig:simu} that the proposed DJL method is very efficient and outperforms all competing methods in almost all cases. We note that the proposed method performs reasonably well even when the sample size is small ($n=50$).
In contrast, kernel-based methods
fail to accurately estimate the value even in some cases when $n=300$. Among the three kernel-based OPE approaches, 
we observe that the method developed by \citet{su2020adaptive} performs better in general. 
A potential limitation of our method is that it takes a longer computational time than the method of \citet{colangelo2020double}. To speed up the dynamic programming algorithm, for instance, the total variation or group-fused-lasso-type penalty can be used as a surrogate of the $L_0$ penalty to reduce the computational complexity \citep[see e.g.,][]{harchaoui2010multiple}. 

\subsection{Real Data: Personalized Dose Finding}\label{sec:real}
Warfarin is commonly used for preventing thrombosis and thromboembolism. 
We use the dataset provided by the International Warfarin Pharmacogenetics \citep{international2009estimation} for analysis. We choose $p = 81$ features considered in \citet{kallus2018policy}. This yields a total of $3964$ with complete records. The outcome is defined as the absolute distance between the international normalized ratio (INR, a measurement of the time it takes for the blood to clot) after the treatment and the ideal value $2.5$, i.e, $Y=-|\hbox{INR}-2.5|$. We use the min-max normalization to convert the range of the dose level $A$ into $[0,1]$. 
To compare among different methods, we calibrate the dataset 
to generate simulated outcomes. This allows us to use simulated data to calculate the bias and variance of each value estimator. Specifically, we first estimate the function $Q(x,a)$ via the MLP regressor using the whole dataset.
The goodness-of-the-fit of the fitted model under the MLP regressor is reported in Table \ref{table:add3} in Appendix \ref{add:exp_res}. 
We next use the fitted function $\widehat{Q}(X,A)$ to simulate the data. For a given sample size $N$, we first randomly sample $N$ feature-treatment pairs $\{(a_j,x_j):1\le j\le N\}$ from 
$\{(A_1,X_1),\cdots,(A_n,X_n)\}$ with replacement. Next, for each $j$, we generate the outcome $y_j$ according to $N\{\widehat{Q}(x_j,a_j),\widehat{\sigma}^2\}$, where $\widehat{\sigma}$ is the standard deviation of the fitted residual $\{Y_i-\widehat{Q}(X_i,A_i)\}_i$. This yields a simulated dataset $\{(x_j,a_j,y_j):1\le j\le N\}$. We are interested in evaluating the mean outcome under the optimal policy as
$
\pi^\star(X) \equiv \argmax_{a \in [0,1]} \widehat{Q}(X,a).
$  


\begin{table}
\centering
	\caption{The absolute error, the standard deviation, and the mean squared error of the estimated values under the optimal policy via  different methods for the Warfarin data. The target value is given by $-0.278$.}\label{table:2} 
		\scalebox{0.88}{\begin{tabular}{llll}
		\toprule
			Methods& Absolute error&Standard deviation&Mean squared error\\
			\midrule
			 Deep Jump Learning & 0.259&0.416 &0.240\\
			\midrule 
			 SLOPE & 0.611&0.755&0.943\\
			\midrule
			 \citet{kallus2018policy}  & 0.662&0.742&0.989\\
			\midrule
			\citet{colangelo2020double}  & 0.442&1.164&1.550\\
			\bottomrule
	\end{tabular}}
\end{table} 
 
We apply the proposed DJL method and the three kernel-based methods to the calibrated Warfarin dataset.
Absolute errors, standard deviations, and mean squared errors of the estimated values under the optimal policy are reported in Table \ref{table:2} over 20 replicates with different evaluation methods.
It can be observed from Table \ref{table:2} that our proposed DJL method achieves much smaller absolute error (0.259) and standard deviation (0.416) than the three kernel-based methods. 
The mean square error of the three competing estimators are at least 3 times larger than DJL. The absolute error and standard deviation of
\citet{kallus2018policy}'s estimator and of the SLOPE in \citet{su2020adaptive} are approximately the same, due to that the bandwidth parameter is optimized. The estimator developed by \citet{colangelo2020double}'s performs the worst. It suffers from a large variance, due to the suboptimal choice of the bandwidth. All these observations are aligned with the findings in our simulation studies.  


\section{Discussion}\label{con}
{We proposed a brand-new OPE algorithm in continuous treatment domains. Combining our theoretical analysis and experiments, we are more confident that our proposed DJL method offers a practically much more useful policy evaluation tool compared to existing kernel-based approaches. There are some potential alternative directions to address the limitation of kernel-based approaches. \citet{majzoubi2020efficient} proposed a tree-based discretization to handle continuous actions in policy optimization for contextual bandits. Extending the tree-based discretization with adaptive pruning in OPE is a possible direction to handle our problem.  Second, our proposed method can be understood as a special local kernel method with the boxcar kernel function, as we adaptively discretize the action space into a set of non-overlapping intervals. It would be practically interesting to investigate how to couple our procedure with general kernel functions. }

 \newpage
\bibliographystyle{agsm}
\bibliography{mycite}
\newpage
\appendix
{\section{Analysis of Computational Complexity of DJL}\label{add:compu_complex}}
We analyze the computational complexity for the proposed method as follows. There are three main dominating parts of the computation: the adaptive discretization, the estimations of conditional mean function and the propensity score function, and the construction of the final value estimator.

First, for the adaptive discretization on the treatment space (the main part of DJL, see Algorithm  \ref{alg1} Part III.3), we use the pruned exact linear time (PELT) method in \citet{killick2012optimal} to solve the dynamic programing. This step requires at least $\mathcal{O}(m)$ computing steps and at most $\mathcal{O}(m^2)$ steps \citep{friedrich2008complexity}. According to Theorem 3.2 in \citet{killick2012optimal}, the expected computational cost is $\mathcal{O}(m)$.

Second, for each step in the linear complexity of adaptive discretization, we need to train the deep neural network for the conditional mean function and the propensity score function to calculate the cost function. Here, the time and space complexity of training a deep learning model varies depending on the actual architecture used. In our implementation, we employ the commonly used multilayer perceptron (MLP) to estimate the function $Q$ and the propensity score in each segment. Suppose we use the standard fully connected MLPs of $w$ width and $d$ depth with feedforward pass and back-propagation under total $e$ epochs. Then according to the complexity analysis of neural networks, the computational complexity of modeling the function $Q$ and the propensity score is $\mathcal{O}\{2 *n e (d-1) w^2 \}$.

For the last part, the construction of the final value estimator based on $\mathcal{L}$-fold cross fitting, which repeats the above two steps $\mathcal{L}$ times. Therefore, by putting the above results together, the total expected computational complexity of the proposed DJL is $\mathcal{O}\{ \mathcal{L} *m *2 *n  e (d-1)w^2\}$. 
Note that the computation for the last part (i.e., cross-fitting) can be easily implemented in parallel computing, and thus the total expected computational complexity of the proposed DJL can be reduced to $\mathcal{O}\{ m *2 * n  e (d-1)w^2\}$.  

\newpage

\section{Additional Experimental Results}\label{add:exp_res}

\begin{figure}[!htp]
	\centering
	\includegraphics[width=.8\textwidth]{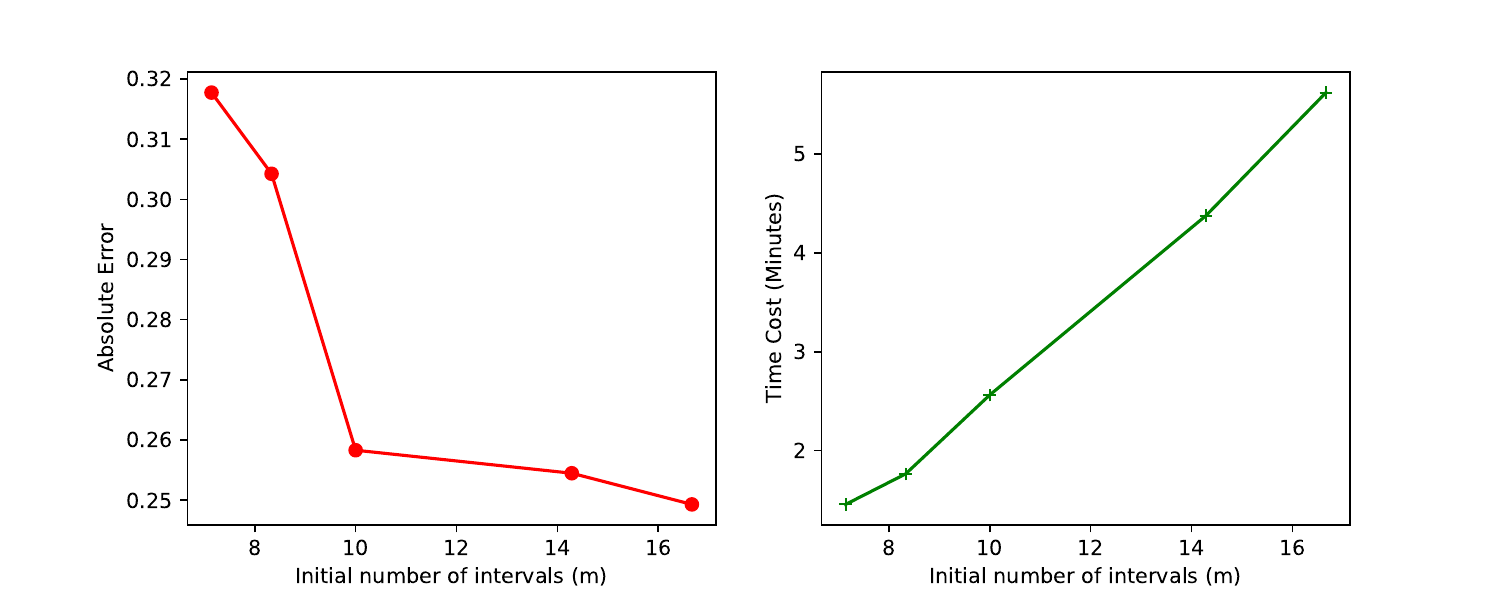}
	\caption{The absolute error of the estimated value and the computational cost (in minutes) under the DJL with different initial number of intervals ($m$) when $n=100$ in Scenario 1.}\label{fig:add1}
\end{figure}

We include additional experimental results in this section. {First, the number of initial intervals $m$ represents a trade-off between the estimation bias and the computational cost, as illustrated in Figure \ref{fig:add1}. In practice, we recommend to set $m = n/10$. When $n$ is small, the performance of the resulting value estimator is not overly sensitive to the choice of $c$ as long as $c$ is not too large. See the left panel of Figure \ref{fig:add1} for details. When $n$ is large, we further investigate the computational capacity of the proposed method by setting $m=n/10$ for large sample sizes and report the corresponding computational time in Table \ref{table:add_rebuttal}. We use Scenario 1 and consider the sample size chosen from $n \in \{1000, 2000, 5000, 10000\}$ for illustration. It turns out that such a choice of $c$ can still handle datasets with a few thousand observations. Here, we use parallel computing to process each fold, as our algorithm employs data splitting and cross-fitting. This largely facilitates the computation, leading to shorter computation time compared to those listed in Table \ref{table:add2}. Finally, when $n$ is extremely large, setting $m=n/10$ might be computationally intensive. 
In addition to parallel computing, there are some other techniques we can use to handle datasets with large sample size. For instance, in the change-point literature, \cite{lu2017intelligent} proposed an intelligence sampling method to identify multiple change points with long time series data. Their method would not lose much statistical efficiency, but is much more computationally efficient. It is possible to adopt such an intelligence sampling method to our setting for adaptive discretization. This would enable our method to handle large datasets.  }

  \begin{table}[!thp]
	\centering
	\caption{The averaged computational cost (in minutes) under the proposed deep jump learning and three kernel-based methods for Scenario 1.}\label{table:add2}
\scalebox{0.9}{
	\begin{tabular}{lllll}
		\toprule
		Methods &Deep Jump Learning&SLOPE  & \citet{kallus2018policy} & \citet{colangelo2020double}  \\
		\midrule
		$n=50$ &$<1$&<1&365&$<1$ \\
		\midrule
		$n=100$ &3&<1&773&$<1$\\
		\midrule
		$n=200$ &7&1&$>1440$ (24 hours)&$<1$\\
		\midrule
		$n=300$ &14&2&$>2880$ (48 hours)&$<1$\\
		\bottomrule
	\end{tabular}}
\end{table} 
\newpage
\begin{table}[!thp]
	\centering
	\caption{The averaged computational cost under the proposed deep jump learning for Scenario 1 with large sample settings.}\label{table:add_rebuttal}
\scalebox{0.9}{
	\begin{tabular}{lllll}
		\toprule
		Sample Size           &             $n=1000$    &        $n=2000$      &      $n=5000$           &$n=10000$\\
		\midrule 
Computational time      &   15.92 minutes   & 30.40 minutes  &  1.32 hours      &   2.86 hours\\
		\bottomrule
	\end{tabular}}
\end{table}

       \begin{table}[!htp]
 \centering
	\caption{The absolute error and the standard deviation (in parentheses) of the estimated values under the optimal policy via the proposed deep jump learning and three kernel-based methods for Scenario 1 to 4.}\label{table:1}
	\scalebox{0.9}{
		\begin{tabular}{llllll}
\toprule
		&$n$ &50 & 100&200&300 \\
		\midrule
		Scenario 1& Deep Jump Learning&0.445(0.381) &0.398(0.391)&0.253(0.269)&0.209(0.210)  \\
		\cmidrule{2-6}
		$V$ =  1.33 & SLOPE&0.392(0.377) &0.385(0.549)&0.329(0.400)&0.344(0.209)  \\
		\cmidrule{2-6}
		&\citet{kallus2018policy} &0.656(0.787) & 0.848(0.799)& 1.163(0.884) &0.537(0.422)\\
		\cmidrule{2-6}
		 & \citet{colangelo2020double} &1.285(1.230) & 1.473(1.304)& 1.826(1.463) &0.934(0.730)\\
		\midrule
		Scenario 2& Deep Jump Learning&0.696(0.376) &0.502(0.311)&0.400(0.219)&0.411(0.168)  \\
		\cmidrule{2-6}
		$V$ =   1.00&  SLOPE&0.620(0.634) &0.859(0.822)&0.749(0.878)&1.209(0.435)  \\
		\cmidrule{2-6}
		&\citet{kallus2018policy} &1.061(1.124)&1.363(1.131)&1.679(1.032) &1.664(0.792)\\
		\cmidrule{2-6}
		 & \citet{colangelo2020double} &1.827(1.371)& 2.292(1.458)& 2.429(1.541) &2.264(1.062) \\
			\midrule
		Scenario 3&Deep Jump Learning& 2.014(0.865)&1.410(0.987)&1.184(0.967)&1.267(0.933)  \\
		\cmidrule{2-6}
		$V$ = 4.86  &  SLOPE&3.660(0.496) &3.185(0.592)&2.897(0.781)&2.037(0.401)  \\
		\cmidrule{2-6}
		& \citet{kallus2018policy} &2.196(2.369)&2.758(2.510)&3.573(2.862) &1.151(1.798)\\
		\cmidrule{2-6}
		 & \citet{colangelo2020double} &2.586(2.825) & 3.172(3.027)&3.949(3.391) &1.367(2.110)\\
		\midrule
		Scenario 4&Deep Jump Learning&0.494(0.485)&0.412(0.426)&0.349(0.383)&0.321(0.315)  \\
	\cmidrule{2-6}
		$V$ = 1.60 &  SLOPE&0.586(0.337) &0.537(0.279)&0.483(0.272)&0.483(0.143)  \\
		\cmidrule{2-6}
		&  \citet{kallus2018policy} &2.192(1.210)&2.740(1.034)&3.354(1.324) &1.555(0.500)\\
	\cmidrule{2-6}
		  & \citet{colangelo2020double} & 2.975(1.789)&3.282(1.525)&3.921(1.927) &1.853(0.751)  \\
		\bottomrule
	\end{tabular}}
\end{table} 

\newpage 

\begin{table}[!htp]
	\centering
	\caption{The averaged size of the final estimated partition ($|\widehat{\mathcal{D}}|$) in comparison to the initial number of intervals ($m$) under the proposed DJL for Scenario 1 to 4.}\label{table:add1}
	\scalebox{0.9}{
	\begin{tabular}{llllll}
		\toprule
		$|\widehat{\mathcal{D}}|$ / $m $ &$n=$50 & $n=$100&$n=$200&$n=$300 \\
		\midrule
		Scenario $1$&  3 / 5 &4 / 10&6 / 20&6 / 30  \\
		\midrule
		Scenario $2$&  4 / 5 &6 / 10&9 / 20&11 / 30  \\
		\midrule
		Scenario $3$&  4 / 5 &6 / 10&8 / 20&10 / 30  \\
		\midrule
		Scenario $4$& 4 / 5 &6 / 10&8 / 20&10 / 30  \\
				\bottomrule
	\end{tabular}}
\end{table}

\begin{table}[!thp]
	\centering
	\caption{The mean squared error (MSE), the normalized root-mean-square-deviation (NRMSD), the mean absolute error (MAE), and the normalized MAE (NMAE) of the fitted model under the multilayer perceptrons regressor, linear regression, and the random forest algorithm, via ten-fold cross-validation.}\label{table:add3}
	\begin{tabular}{cccc}
		\hline
		\hline
		Method &Multilayer Perceptrons Regressor& Linear Regression & Random Forest \\
		\hline
		\hline
		MSE&0.06&0.09&0.08\\
		\hline
		NRMSD&0.13&0.16&0.15\\
		\hline
		MAE&0.19&0.23&0.22\\
		\hline
		NMAE&0.10&0.12&0.12\\
		\hline
	\end{tabular}
\end{table}    \footnote{$MSE = {1\over n}\sum_{i=1}^n (Y_i -\widehat{Y}_i)^2$. See \url{https://en.wikipedia.org/wiki/Mean_squared_error.}\\
$NRMSD = {\sqrt{MSE}\over \max(Y)- \min(Y)}$. See \url{https://en.wikipedia.org/wiki/Root-mean-square_deviation.} \\
$MAE = {1\over n}\sum_{i=1}^n |Y_i -\widehat{Y}_i|$. See \url{https://en.wikipedia.org/wiki/Mean_absolute_error.} \\
$NMAE =  {MAE\over \max(Y)- \min(Y)}$. See \url{https://en.wikipedia.org/wiki/Root-mean-square_deviation.} 
}

\section{Rate of Convergence of Kernel-Based Estimators}\label{eqn:ratekernel}
\subsection{Convergence Rate under Model 1}
Consider the following piecewise constant function $Q$ 
\begin{eqnarray*}
	Q(x,a)=\left\{\begin{array}{ll}
		0, & \hbox{if}~a\le 1/2,\\
		1, & \hbox{otherwise}.
	\end{array}
	\right.
\end{eqnarray*}
Define a policy $\pi$ such that the density function of $\pi(X)$ equals
\begin{eqnarray*}
	\left\{\begin{array}{ll}
		4/3, &\hbox{if}~1/4\le \pi(x)\le 1/2,\\
		2/3, &\hbox{else~if~}1/2\le \pi(x)<4/3,\\
		0, &\hbox{otherwise}.
	\end{array}
	\right.
\end{eqnarray*}
We aim to show for such $Q$ and $\pi$, the best possible convergence rate of kernel-based estimator is $n^{-1/3}$. 

We first consider its variance. Suppose the conditional variance of $Y|A,X$ is uniformly bounded away from $0$. Similar to Theorem 1 of \cite{colangelo2020double}, we can show the variance of kernel based estimator is lower bounded by $O(1) (nh)^{-1}$ where $O(1)$ denotes some positive constant. 

We next consider its bias. Since the behavior policy is known, the bias is equal to 
\begin{eqnarray*}
	\Mean \left(\frac{K[\{A-\pi(X)\}/h]}{h b(A|X)}[Y-Q\{X,\pi(X)\}]\right)=\Mean \left(\frac{K[\{A-\pi(X)\}/h]}{h b(A|X)}[Q(X,A)-Q\{X,\pi(X)\}]\right)\\=\Mean  \left(\int_{\pi(X)-h/2}^{\pi(X)+h/2} K\left\{\frac{a-\pi(X)}{h}\right\}[\mathbb{I}\{\pi(X)\le 1/2<a\}-\mathbb{I}\{a\le 1/2<\pi(X)\}]da \right).
\end{eqnarray*}
Using the change of variable $a=ht+\pi(X)$, the bias equals
\begin{eqnarray*}
	\Mean \left(\int_{-1/2}^{1/2} K(t)[\mathbb{I}\{\pi(X)\le 1/2<\pi(X)+ht\}-\mathbb{I}\{\pi(X)+ht\le 1/2<\pi(X)\}]dt \right).
\end{eqnarray*}
Consider any $0<h\le \epsilon$ for some sufficiently small $\epsilon>0$. The bias is then equal to
\begin{eqnarray*}
	&&\frac{4}{3}\int_{1/2-\epsilon/2}^{1/2}\int_{-1/2}^{1/2} K(t)\{\mathbb{I}(a\le 1/2<a+ht)-\mathbb{I}(a+ht\le 1/2<a)\}dtda\\
	&+&\frac{2}{3}\int_{1/2}^{1/2+\epsilon/2}\int_{-1/2}^{1/2} K(t)\{\mathbb{I}(a\le 1/2<a+ht)-\mathbb{I}(a+ht\le 1/2<a)\}dtda.
\end{eqnarray*}
Under the symmetric condition on the kernel function, the above quantity is equal to
\begin{eqnarray*}
	\frac{2}{3}\int_{1/2-h/2}^{1/2} \int_{(1-2a)/2h}^{1/2} K(t)dtda\ge \frac{2}{3}\int_{1/2-h/2}^{1/2-h/4} \int_{(1-2a)/2h}^{1/2} K(t)dtda\\
	\ge \frac{2}{3}\int_{1/2-h/2}^{1/2-h/4} \int_{1/4}^{1/2} K(t)dtda=\frac{h}{6} \int_{1/4}^{1/2} K(t)dt.
\end{eqnarray*}
Consequently, the bias is lower bounded by $O(1) h$ where $O(1)$ denotes some positive constant. 

To summarize, the root mean squared error of kernel based estimator is lower bounded by $O(1) \{(nh)^{-1/2}+h\}$ where $O(1)$ denotes some positive constant. The optimal choice of $h$ that minimizes such lower bound would be of the order $n^{-1/3}$. Consequently, the convergence rate is lower bounded by $O(1) n^{-1/3}$. 
\subsection{Convergence Rate under Model 2}
Similar to the case under Model 1, we can show the variance of kernel-based estimator is lower bounded by $O(n^{-1} h^{-1})$ in cases where the conditional variance of $Y$ given $(A,X)$ is uniformly bounded away from zero. 

Consider the conditional mean function $Q$ 
\begin{eqnarray*}
	Q(x,a)=C h^{-1} K\left\{\frac{a-\pi(x)}{h}\right\},
\end{eqnarray*}
for some constant $C>0$. We aim to derive the bias of kernel-based estimator under such a choice of the conditional mean function $Q$. Using similar arguments in the case where Model 1 holds, we can show the bias equals
\begin{eqnarray*}
	\Mean \left(C^{-1} \frac{K^2[\{A-\pi(X)\}/h]}{h^2 b(A|X)} \right)\ge C^{-1}\Mean \left( \frac{K^2[\{A-\pi(X)\}/h]}{h^2} \right). 
\end{eqnarray*}
Similarly, we can show the right-hand-side is lower bounded by $O(1) h$. This implies that the convergence rate is at least $O(1) (n^{-1}h^{-1}+h)$ under Model 2.

\section{Technical Proof}\label{tech_proof}
Throughout the proof, we use $c,C,c_0,\bar{c},c_*$, etc., to denote some universal constants whose values are allowed to change from place to place. Let $O_i=\{X_i,Y_i\}$ denote the data summarized from the $i$th observation. For any two positive sequences $\{a_n\}_n$ and $\{b_n\}_n$. The notation $a_n\asymp b_n$ means that there exists some universal constant $c>1$ such that $c^{-1}b_n\le a_n\le cb_n$ for any $n$. The notation $a_n\propto b_n$ means that there exists some universal constant $c>0$ such that $a_n\le cb_n$ for all $n$. 

Proofs of Theorems \ref{thm1} and \ref{thm2} rely on Lemmas \ref{lemma0}, \ref{lemma1} and \ref{lemma2}. In particular, Lemma \ref{lemma0} establishes the uniform convergence rate of $\widehat{q}_{\mathcal{I}}^{(\ell)}$ for any $\mathcal{I}$ whose length is no shorter than $o(\gamma_n)$ and belongs to the set of intervals:
\begin{eqnarray*}
	\mathfrak{I}(m)&=&\{ [i_1/m,i_2/m):\hbox{for~some~integers}~i_1\hbox{~and~}i_2~\hbox{that~satisfy}~0\le i_1<i_2<m\}\\
	&\cup& \{ [i_3/m,1]:\hbox{for~some~integers}~i_3~\hbox{that~satisfy}~0\le i_3<m\}.
\end{eqnarray*}
To state this lemma, we first introduce some notations. For any such interval $\mathcal{I}$, define the function $q_{\mathcal{I},0}(x)=\Mean (Y|A\in \mathcal{I},X=x)$. 
It is immediate to see that the definition of $q_{\mathcal{I},0}$ here is consistent with the one defined in \eqref{step} for any $\mathcal{I}\subseteq \mathcal{D}_0$. 
\begin{lemma}\label{lemma0}
	Assume either conditions in Theorem 1 or 2 are satisfied. Then there exists some constant $\bar{C}>0$ such that  the following holds with probability at least $1-O(n^{-2})$: For any $1\le \ell\le \mathcal{L}$, $ \mathcal{I}\in \mathfrak{I}(m)$ and $|\mathcal{I}|\ge c\gamma_n $,
	\begin{eqnarray}\label{eqn:event0}
	\Mean  [|q_{\mathcal{I},0}(X)-\widehat{q}_{\mathcal{I}}^{(\ell)}(X)|^2\{O_i\}_{i\in \mathbb{L}_{\ell}^c}] \le \bar{C} (n|\mathcal{I}|)^{-2\beta/(2\beta+p)}\log^8 n.
	\end{eqnarray}
\end{lemma}
Here, the expectation in \eqref{eqn:event0} is taken with respect to a testing sample $X$.

\begin{lemma}\label{lemma1} 
	Assume either conditions in Theorem 1 or 2 are satisfied. Then there exists some constant $\bar{C}>0$ such that the followings hold with probability at least $1-O(n^{-2})$: For any $1\le \ell\le \mathcal{L}$, $\mathcal{I}\in \mathfrak{I}(m)$ and $|\mathcal{I}|\ge c\gamma_n$,
	\begin{eqnarray*}
		\sum_{\substack{\mathcal{I}\in \widehat{\mathcal{D}}^{(\ell)} }}\left|\sum_{i\in \mathbb{L}_{\ell}^c} \mathbb{I}(A_i\in \mathcal{I})\{Y_i-q_{\mathcal{I},0}(X_i)\}  \{\widehat{q}_{\mathcal{I}}^{(\ell)}(X_i)-q_{\mathcal{I},0}(X_i)\} \right|\le \bar{C} (n|\mathcal{I}|)^{p/(2\beta+p)}log^8 n.
	\end{eqnarray*}
\end{lemma}

\begin{lemma}\label{lemma2} 
	Assume either conditions in Theorem 1 or 2 are satisfied. Then the following events occur with probability at least $1-O(n^{-2})$: there exists some constant $c>0$ such that $\min_{\mathcal{I}\in \widehat{\mathcal{D}}^{(\ell)}}|\mathcal{I}|\ge c \gamma_n$ for any $1\le \ell \le \mathcal{L}$. 
\end{lemma}
We first present the proofs for these three lemmas. Next we present the proofs for Theorems \ref{thm1} and \ref{thm2}. 

\subsection{Proof of Lemma \ref{lemma0}}\label{sec:prooflemma0} 
The number of folds $\mathcal{L}$ is bounded. It suffices to derive the uniform convergence rate for each $\ell$. By definition, $\widehat{q}_{\mathcal{I}}^{(\ell)}$ is the minimizer of the least square loss, $\argmin_{q\in \mathcal{Q}_{\mathcal{I}}}\sum_{i\in \mathbb{L}_{\ell}^c} \mathbb{I}(A_i\in \mathcal{I}) |Y_i-q(X_i)|^2$. It follows that
\begin{eqnarray*}
	\sum_{i\in \mathbb{L}_{\ell}^c} \mathbb{I}(A_i\in \mathcal{I}) |Y_i-\widehat{q}_{\mathcal{I}}^{(\ell)}(X_i)|^2\le \sum_{i\in \mathbb{L}_{\ell}^c} \mathbb{I}(A_i\in \mathcal{I}) |Y_i-q(X_i)|^2,
\end{eqnarray*}
for all $q\in \mathcal{Q}_{\mathcal{I}}$. Recall that $q_{\mathcal{I},0}(x)=\Mean (Y|A\in \mathcal{I},X=x)$, we have $\Mean [\mathbb{I}(A\in \mathcal{I})\{Y-q_{\mathcal{I},0}(X)\}|X]=0$. A simple calculation yields 
\begin{eqnarray*}
	\sum_{i\in \mathbb{L}_{\ell}^c} \mathbb{I}(A_i\in \mathcal{I}) |q_{\mathcal{I},0}(X_i)-\widehat{q}_{\mathcal{I}}^{(\ell)}(X_i)|^2\le \sum_{i\in \mathbb{L}_{\ell}^c} \mathbb{I}(A_i\in \mathcal{I}) |q_{\mathcal{I},0}(X_i)-q(X_i)|^2\\+2\sum_{i\in \mathbb{L}_{\ell}^c} \mathbb{I}(A_i\in \mathcal{I})\{Y_i-q_{\mathcal{I},0}(X_i)\}\{\widehat{q}_{\mathcal{I}}^{(\ell)}(X_i)-q_{\mathcal{I},0}(X_i)\},
\end{eqnarray*} 
for any $q$ and $\mathcal{I}$. 

The first term on the right-hand-side measures the approximation bias of the class of deep neural networks. Since $\Mean [\mathbb{I}(A\in \mathcal{I})\{Y-q_{\mathcal{I},0}(X)\}|X]=0$, the second term corresponds to the stochastic error. The rest of the proof is divided into three parts. In Part 1, we bound the approximation error. In Part 2, we bound the stochastic error. Finally, we combine these two parts together to derive the uniform convergence rate for $\widehat{q}_{\mathcal{I}}^{(\ell)}$.

\smallskip

\noindent \textit{Part 1}. Under the given condition, we have $Q(\bullet,a)\in \Phi(\beta,c)$, $b(a|\bullet)\in \Phi(\beta,c)$ for some $c>0$ and any $a$. We now argue that there exists some constant $C>0$ such that $q_{\mathcal{I},0}\in \Phi(\beta,C)$ for any $\mathcal{I}$. This can be proven based on the relation that
\begin{eqnarray*}
	q_{\mathcal{I},0}(x)=\frac{\int_{\mathcal{I}} Q(x,a) b(a|x)da}{\int_{\mathcal{I}} b(a|x)da}.
\end{eqnarray*}
Specifically, we have that $\sup_x |q_{\mathcal{I},0}(x)|\le \sup_{a,x} |Q(x,a)|\le c$. Suppose $\beta\le 1$. For any $x_1,x_2\in \mathcal{X}$, consider the difference $|q_{\mathcal{I},0}(x_1)-q_{\mathcal{I},0}(x_2)|$. Under the positivity assumption, we have $\inf_{a,x}b(a|x)\ge c_*$ for some $c_*>0$. It follows that
\begin{eqnarray*}
	|q_{\mathcal{I},0}(x_1)-q_{\mathcal{I},0}(x_2)|\le \frac{\int_{\mathcal{I}} |Q(x_1,a)-Q(x_2,a)| b(a|x_1)da}{\int_{\mathcal{I}} b(a|x_1)da}\\+\frac{\int_{\mathcal{I}} |Q(x_2,a)||b(a|x_1)-b(a|x_2)|da }{\int_{\mathcal{I}} b(a|x_1)da}
	+\frac{\int_{\mathcal{I}}|Q(x_2,a)|b(a|x_2)da \int_{\mathcal{I}}|b(a|x_1)-b(a|x_2)|da }{\int_{\mathcal{I}} b(a|x_1)da \int_{\mathcal{I}} b(a|x_2)da}\\
	\le c\|x_1-x_2\|^{\beta-\floor{\beta}}+2\frac{c^2}{c_*} \|x_1-x_2\|^{\beta-\floor{\beta}}.
\end{eqnarray*}
Consequently, $q_{\mathcal{I},0} \in \Phi(\beta, c+2c^2/c_*^2)$. 

Suppose $\beta>1$. Then both $Q(\bullet,a)$ and $b(a|\bullet)$ are $\floor{\beta}$-differentiable. By changing the order of integration and differentiation, we can show that $q_{\mathcal{I},0}(x)$ is $\floor{\beta}$-differentiable as well. As an illustration, when $\beta<2$, we have $\floor{\beta}=1$. According to the chain rule, we have
\begin{eqnarray*}
	\frac{\partial q_{\mathcal{I},0}(x)}{\partial x^j}=\frac{\int_{\mathcal{I}} \{\partial Q(x,a)/\partial x^j\} b(a|x)da}{\int_{\mathcal{I}} b(a|x)da}+\frac{\int_{\mathcal{I}}Q(a|x) \{\partial b(a|x)/\partial x^j\} da}{\int_{\mathcal{I}} b(a|x)da}\\
	-\frac{\int_{\mathcal{I}}Q(a|x) b(a|x)da \int_{\mathcal{I}} \{\partial b(a|x)/\partial x^j\}da}{\{\int_{\mathcal{I}} b(a|x)da\}^2}.
\end{eqnarray*}
Moreover, using similar arguments in proving $q_{\mathcal{I},0} \in \Phi(\beta, c+2c^2/c_*^2)$ when $\beta<1$, we can show that all the partial derivatives of $q_{\mathcal{I},0}(x)$ up to the $\floor{\beta}$th order are uniformly bounded for all $\mathcal{I}$. In addition, all the $\floor{\beta}$th order partial derivatives are H{\"o}lder continuous with exponent $\beta-\floor{\beta}$. This implies that $q_{\mathcal{I},0}\in \Phi(\beta,C)$ for some constant $C>0$ and any $\mathcal{I}$. 

It is shown in Lemma 7 of \cite{farrell2021deep} that for any $\epsilon>0$, there exists a deep neural network architecture that approximates $q_{\mathcal{I},0}$ with the uniform approximation error upper bounded by $\epsilon$, and satisfies $W_{\mathcal{I}}\le \bar{C} \epsilon^{-p/\beta}(\log \epsilon^{-1}+1)$ and $L_{\mathcal{I}}\le \bar{C} (\log \epsilon^{-1}+1)$ for some constant $\bar{C}>0$. These upper bounds will be used later in Part 2. The detailed value of $\epsilon$ will be specified below. It follows that for any $\mathcal{I}$, the bias term can be upper bounded by
\begin{eqnarray}\label{eqn:lemma01}
	\sum_{i\in \mathbb{L}_{\ell}^c} \mathbb{I}(A_i\in \mathcal{I}) |q_{\mathcal{I},0}(X_i)-q(X_i)|^2\le \epsilon^2 \sum_{i\in \mathbb{L}_{\ell}^c} \mathbb{I}(A_i\in \mathcal{I}).
\end{eqnarray}
We next provide an upper bound for the right-hand-side. 
Since $A$ has a bounded probability density function, the variance $\Var\{\mathbb{I}(A_i\in \mathcal{I})\}$ is upper bounded by $\sqrt{\Mean \mathbb{I}(A_i\in \mathcal{I})}\le \bar{c}\sqrt{|\mathcal{I}|}$ for some universal constant $\bar{c}>0$. It follows from Bernstein's inequality that
\begin{eqnarray*}
	\prob\left\{\sum_{i\in \mathbb{L}_{\ell}^c} \mathbb{I}(A_i\in \mathcal{I})-|\mathbb{L}_{\ell}^c|\Mean \mathbb{I}(A\in \mathcal{I})\ge t \right\}\le \exp\left(-\frac{t^2/2}{\bar{c}^2 |\mathbb{L}_{\ell}^c||\mathcal{I}|+t/3}\right),
\end{eqnarray*}
for any $t$ and $\mathcal{I}$. Set $t_{\mathcal{I}}=6\max(\bar{c}\sqrt{n|\mathcal{I}|\log n},|\mathcal{I}|\log n)$, the right-hand-side is upper bounded by $n^{-4}$. Since $m\asymp n$ and the number of intervals $\mathcal{I}$ in $\mathfrak{I}(m)$ is upper bounded by $m^2$, it follows from Bonferroni's inequality that
\begin{eqnarray*}
	\prob\left[\bigcup_{\mathcal{I}\in \mathfrak{I}(m)}\left\{\sum_{i\in \mathbb{L}_{\ell}^c} \mathbb{I}(A_i\in \mathcal{I})-|\mathbb{L}_{\ell}^c|\Mean \mathbb{I}(A\in \mathcal{I})\ge t_{\mathcal{I}}\right\} \right]\le m^2n^{-4}=O(n^{-2}).
\end{eqnarray*}
As such, with probability at least $1-O(n^{-2})$, we have that $\sum_{i\in \mathbb{L}_{\ell}^c} \mathbb{I}(A_i\in \mathcal{I})-|\mathbb{L}_{\ell}^c|\Mean \mathbb{I}(A\in \mathcal{I})\le t_{\mathcal{I}}$
uniformly for all $\mathcal{I}$, or equivalently, $\sum_{i\in \mathbb{L}_{\ell}^c} \mathbb{I}(A_i\in \mathcal{I})\le |\mathbb{L}_{\ell}^c|\bar{c}|\mathcal{I}|+t_{\mathcal{I}}$. Consider a subset of intervals $\mathcal{I}$ with $|\mathcal{I}|\ge c\gamma_n$ for any constant $c>0$. Under the given conditions on $\gamma_n$, we have 
\begin{eqnarray}\label{eqne0}
	\sum_{i\in \mathbb{L}_{\ell}^c} \mathbb{I}(A_i\in \mathcal{I})\le n\bar{c}^*|\mathcal{I}|,\,\,\,\text{ for any } \mathcal{I}\,\,\text{ such that }\,\,|\mathcal{I}|\ge c\gamma_n,
\end{eqnarray}
for some constant $\bar{c}^*>0$. It follows from \eqref{eqn:lemma01} that the following holds with probability at least $1-O(n^{-2})$: for any $\mathcal{I}\in\mathfrak{I}(m)$ such that $|\mathcal{I}|\ge c\gamma_n$, we have
\begin{eqnarray*}
	\sum_{i\in \mathbb{L}_{\ell}^c} \mathbb{I}(A_i\in \mathcal{I}) |q_{\mathcal{I},0}(X_i)-q(X_i)|^2\le \bar{c}^*\epsilon^2 n|\mathcal{I}|.
\end{eqnarray*}
Set $\epsilon$ to $(n|\mathcal{I}|)^{-\beta/(2\beta+p)}$, it follows that
\begin{eqnarray}\label{eqn:bias}
\sum_{i\in \mathbb{L}_{\ell}^c} \mathbb{I}(A_i\in \mathcal{I}) |q_{\mathcal{I},0}(X_i)-q(X_i)|^2\le \bar{c}^*(n|\mathcal{I}|)^{-2\beta/(2\beta+p)} (n|\mathcal{I}|).
\end{eqnarray}
$W_{\mathcal{I}}$ and $L_{\mathcal{I}}$ are upper bounded by $\bar{C} (n|\mathcal{I}|)^{p/(2\beta+p)} (\beta\log (n|\mathcal{I}|)/(2\beta+p)+1)$ and $\bar{C}(\beta \log (n|\mathcal{I}|)/(2\beta+p)+1)$, respectively. This completes the proof for Part 1. 

\smallskip

\noindent \textit{Part 2}. For the function class of deep neural networks $Q_{\mathcal{I}}$, we use $\theta_{\mathcal{I}}$ to denote the parameters in deep neural networks. This allows us to represent $\mathcal{Q}_{\mathcal{I}}$ as $\{q_{\mathcal{I}}(\bullet,\theta_{\mathcal{I}}):\theta_{\mathcal{I}}\}$ We will apply the empirical process theory \citep[see e.g.,][]{van1996weak} to bound the stochastic error. Let $\widehat{\theta}_{\mathcal{I}}$ be the estimated parameter in $\widehat{q}_{\mathcal{I}}^{(\ell)}$. Define 
\begin{eqnarray*}
	\sigma^2(\mathcal{I},\theta)=\Mean\left\{ \mathbb{I}(A\in \mathcal{I})|q_{\mathcal{I},0}(X)-q_{\mathcal{I},0}(X,\theta)|^2\right\},
\end{eqnarray*}
for any $\theta$ and $\mathcal{I}$. Consider two separate cases, corresponding to $\sigma(\mathcal{I},\widehat{\theta}_{\mathcal{I}})\le |\mathcal{I}|^{1/2}(n|\mathcal{I}|)^{-\beta/(2\beta+p)}$ and $\sigma(\mathcal{I},\widehat{\theta}_{\mathcal{I}})> |\mathcal{I}|^{1/2}(n|\mathcal{I}|)^{-\beta/(2\beta+p)}$, respectively. We focus our attentions on the latter class of intervals. In Part 3, we will show that for those intervals,  \begin{eqnarray*}\sigma(\mathcal{I},\widehat{\theta}_{\mathcal{I}})\le O(1) |\mathcal{I}|^{1/2}(n|\mathcal{I}|)^{-\beta/(2\beta+p)} \log^4 n,\end{eqnarray*} for some universal constant $O(1)$. This implies that for any $\mathcal{I}$, we have
\begin{eqnarray}\label{eqn:sigmaI}
\sigma(\mathcal{I},\widehat{\theta}_{\mathcal{I}})\le O(1) |\mathcal{I}|^{1/2}(n|\mathcal{I}|)^{-\beta/(2\beta+p)} \log^4 n. 
\end{eqnarray}

We consider bounding a scaled version of the stochastic error,
\begin{eqnarray*}
	\frac{1}{\sigma(\mathcal{I},\widehat{\theta}_{\mathcal{I}})}\sum_{i\in \mathbb{L}_{\ell}^c} \mathbb{I}(A_i\in \mathcal{I})\{Y_i-q_{\mathcal{I},0}(X_i)\}\{\widehat{q}_{\mathcal{I}}^{(\ell)}(X_i)-q_{\mathcal{I},0}(X_i)\}.
\end{eqnarray*}
Its absolute value can be upper bounded by
\begin{eqnarray*}
	\mathbb{Z}(\mathcal{I})\equiv \sup_{\theta} \left| \frac{1}{\sigma(\mathcal{I},\theta)}\sum_{i\in \mathbb{L}_{\ell}^c} \mathbb{I}(A_i\in \mathcal{I})\{Y_i-q_{\mathcal{I},0}(X_i)\}\{q_{\mathcal{I},0}(X_i,\theta)-q_{\mathcal{I},0}(X_i)\} \right|,
\end{eqnarray*}
where the supremum is taken over all $\theta$ such that $\sigma(\mathcal{I},\theta)>|\mathcal{I}|^{1/2} (n|\mathcal{I}|)^{-\beta/(2\beta+p)}$. 

For a given $\theta$, the empirical sum has zero mean.  Under the boundedness assumption on $Y$, its variance is upper bounded by some universal constant. In addition, each quantity $\sigma^{-1}(\mathcal{I},\theta) \mathbb{I}(A_i\in \mathcal{I})\{Y_i-q_{\mathcal{I},0}(X_i)\}\{q_{\mathcal{I},0}(X_i,\theta)-q_{\mathcal{I},0}(X_i)\}$ is upper bounded by $O(1) |\mathcal{I}|^{-1/2} (n|\mathcal{I}|)^{\beta/(2\beta+p)}$ for some universal constant $O(1)$. This allows us to apply the tail inequality developed by \cite{massart2000constants} to bounded the empirical process. See also Theorem 2 of \cite{adamczak2008tail}.  Specifically, for all $t>0$ and $\mathcal{I}$ that satisfies $\sigma(\mathcal{I},\widehat{\theta}_{\mathcal{I}})> |\mathcal{I}|^{1/2}(n|\mathcal{I}|)^{-\beta/(2\beta+p)}$, we obtain with probability at least $1-\exp(t)$ that 
\begin{eqnarray}\label{eqn:event}
\mathbb{Z}(\mathcal{I})\le 2 \Mean \mathbb{Z}(\mathcal{I})+\bar{c}\sqrt{tn}+t\bar{c}|\mathcal{I}|^{-1/2}(n|\mathcal{I}|)^{\beta/(2\beta+p)},
\end{eqnarray}
for some constant $\bar{c}>0$. By setting $t=3\log n$, the probability $1-\exp(t)=1-n^{-3}$. Notice that the number of intervals $\mathcal{I}$ is upper bounded by $O(n^2)$, under the condition that $m$ is proportional to $n$. By Bonferroni's inequality, we obtain that \eqref{eqn:event} holds with probability at least $1-O(n^{-2})$ for any $\mathcal{I}$. Under the given condition on $\gamma_n$, for any interval $\mathcal{I}$ such that $|\mathcal{I}|\ge c\gamma_n$, the last term on the right-hand-side of \eqref{eqn:event} is $o(\sqrt{n})$. It follows that the following occurs with probability $1-O(n^{-2})$,
\begin{eqnarray}\label{eqn:event2}
\mathbb{Z}(\mathcal{I})\le 2 \Mean \mathbb{Z}(\mathcal{I})+2\bar{c}\sqrt{n\log n},
\end{eqnarray}
for all $\mathcal{I}$ such that $|\mathcal{I}|\ge c\gamma_n$ and $\sigma(\mathcal{I},\widehat{\theta}_{\mathcal{I}})> |\mathcal{I}|^{1/2}(n|\mathcal{I}|)^{-\beta/(2\beta+p)}$. 

We next provide an upper bound for $\Mean \mathbb{Z}(\mathcal{I})$. Toward that end, we will apply the maximal inequality developed in Corollary 5.1 of \cite{chernozhukov2014gaussian}. We first observe that the class of empirical sum indexed by $\theta$ belongs to the VC subgraph class with VC-index upper bounded by $O(W_{\mathcal{I}} L_{\mathcal{I}}\log (W_{\mathcal{I}}))$. It follows that for any $\mathcal{I}$ such that $|\mathcal{I}|\ge c\gamma_n$,  $\sigma(\mathcal{I},\widehat{\theta}_{\mathcal{I}})> |\mathcal{I}|^{1/2}(n|\mathcal{I}|)^{-\beta/(2\beta+p)}$, 
\begin{eqnarray*}
	\Mean \mathbb{Z}(\mathcal{I})\propto \sqrt{nW_{\mathcal{I}}L_{\mathcal{I}}\log (W_{\mathcal{I}})\log n}+W_{\mathcal{I}}L_{\mathcal{I}}\log (W_{\mathcal{I}})\log n.
\end{eqnarray*}
Based on the upper bounds on $W_{\mathcal{I}}$ and $L_{\mathcal{I}}$ developed in Part 1, the right-hand-side is upper bounded by
\begin{eqnarray*}
	O(1) (n|\mathcal{I}|)^{p/(4\beta+2p)}\sqrt{n\log^4 n}+O(1) |\mathcal{I}|^{-1/2} (n|\mathcal{I}|)^{p/(2\beta+p)} \log^4 n,
\end{eqnarray*}
where $O(1)$ denotes some universal constant. It is of the order $O\{ n^{1/2}(n|\mathcal{I}|)^{p/(4\beta+2p)}\log^4 n \}$. This yields that
\begin{eqnarray*}
	\Mean \mathbb{Z}(\mathcal{I})\propto n^{1/2}(n|\mathcal{I}|)^{p/(4\beta+2p)}\log^4 n.
\end{eqnarray*}
This together with \eqref{eqn:event} and \eqref{eqn:event2} yields that with probability at least $1-O(n^{-2})$, the scaled stochastic error is upper bounded by $n^{1/2}(n|\mathcal{I}|)^{p/(4\beta+2p)}\log^4 n$. As such, with probability at least $1-O(n^{-2})$, we obtain that
\begin{eqnarray*}
	\left|\sum_{i\in \mathbb{L}_{\ell}^c} \mathbb{I}(A_i\in \mathcal{I})\{Y_i-q_{\mathcal{I},0}(X_i)\}\{\widehat{q}_{\mathcal{I}}^{(\ell)}(X_i)-q_{\mathcal{I},0}(X_i)\}\right|\propto \sigma(\mathcal{I},\widehat{\theta}_{\mathcal{I}}) n^{1/2}(n|\mathcal{I}|)^{p/(4\beta+2p)}\log^4 n,
\end{eqnarray*}
for any $\mathcal{I}$ such that $|\mathcal{I}|\ge c\gamma_n$,  $\sigma(\mathcal{I},\widehat{\theta}_{\mathcal{I}})> |\mathcal{I}|^{1/2}(n|\mathcal{I}|)^{-\beta/(2\beta+p)}$. 
By Cauchy-Schwarz inequality, the left-hand-side can be further upper bounded by
\begin{eqnarray*}
	\frac{n\sigma^2(\mathcal{I},\widehat{\theta}_{\mathcal{I}})}{4}+ O(1)  (n|\mathcal{I}|)^{p/(2\beta+p)}\log^8 n,
\end{eqnarray*}
where $O(1)$ denotes some universal positive constant. This completes the proof for Part 2.

\smallskip

\noindent \textit{Part 3}. Combining the results in Part 1 and Part 2, we obtain that for any $\mathcal{I}$ such that $|\mathcal{I}|\ge c\gamma_n$,  $\sigma(\mathcal{I},\widehat{\theta}_{\mathcal{I}})> |\mathcal{I}|^{1/2}(n|\mathcal{I}|)^{-\beta/(2\beta+p)}$, 
\begin{eqnarray*}
	\sum_{i\in \mathbb{L}_{\ell}^c} \mathbb{I}(A_i\in \mathcal{I}) |q_{\mathcal{I},0}(X_i)-\widehat{q}_{\mathcal{I}}^{(\ell)}(X_i)|^2\le \frac{n\sigma^2(\mathcal{I},\widehat{\theta}_{\mathcal{I}})}{4}+ O(1)  (n|\mathcal{I}|)^{p/(2\beta+p)}\log^8 n,
\end{eqnarray*}
with probability at least $1-O(n^{-2})$. As for the left-hand-side, we notice that
\begin{eqnarray*}
	&&\sum_{i\in \mathbb{L}_{\ell}^c} \mathbb{I}(A_i\in \mathcal{I}) |q_{\mathcal{I},0}(X_i)-\widehat{q}_{\mathcal{I}}^{(\ell)}(X_i)|^2\\
	\ge&& |\mathbb{L}_{\ell}^c|\sigma^2(\mathcal{I},\widehat{\theta}_{\mathcal{I}})-
	\left|\sum_{i\in \mathbb{L}_{\ell}^c} \mathbb{I}(A_i\in \mathcal{I}) |q_{\mathcal{I},0}(X_i)-\widehat{q}_{\mathcal{I}}^{(\ell)}(X_i)|^2-|\mathbb{L}_{\ell}^c|\sigma^2(\mathcal{I},\widehat{\theta}_{\mathcal{I}})\right|.
\end{eqnarray*}
Using similar arguments in Part 2, we can show that the second line is upper bounded by $n\sigma^2(\mathcal{I},\widehat{\theta}_{\mathcal{I}})/8+ O(1)  (n|\mathcal{I}|)^{p/(2\beta+p)}\log^8 n$, with probability at least $1-O(n^{-2})$, for any $\mathcal{I}$ such that $|\mathcal{I}|\ge c\gamma_n$,  $\sigma(\mathcal{I},\widehat{\theta}_{\mathcal{I}})> |\mathcal{I}|^{1/2}(n|\mathcal{I}|)^{-\beta/(2\beta+p)}$. Since $\mathbb{L}_{\ell}^c\ge n/2$, we obtain
\begin{eqnarray*}
	\left(\frac{1}{2}-\frac{1}{4}-\frac{1}{8}\right)\sigma^2(\mathcal{I},\widehat{\theta}_{\mathcal{I}})=\frac{1}{8}\sigma^2(\mathcal{I},\widehat{\theta}_{\mathcal{I}})\propto (n|\mathcal{I}|)^{-2\beta/(2\beta+p)}\log^8 n.
\end{eqnarray*}
This yields the desired uniform upper bound for $\sigma^2(\mathcal{I},\widehat{\theta}_{\mathcal{I}})$. We thus obtain \eqref{eqn:sigmaI} holds with probability at least $1-O(n^{-2})$. 

Under the assumption that the density function $b(a|x)$ is uniformly bounded away from zero, we obtain 
\begin{eqnarray*}
	\sigma^2(\mathcal{I},\widehat{\theta}_{\mathcal{I}})\le c|\mathcal{I}| \Mean |q_{\mathcal{I},0}(X)-\widehat{q}_{\mathcal{I}}^{(\ell)}(X)|^2,
\end{eqnarray*}
for some constant $c>0$. This assertion thus follows. 


\subsection{Proof of Lemma \ref{lemma1}} 
The assertion can be proven in a similar manner as Part 2 of the proof of Lemma \ref{lemma0}. We omit the details to save space.

\subsection{Proof of Lemma \ref{lemma2}}
Consider a given interval $\mathcal{I}\in \widehat{\mathcal{D}}^{(\ell)}$. Suppose $|\mathcal{I}|<c \gamma_n$. The value of the constant $c$ will be determined later. Then, for sufficiently large $n$, we can find some interval $\mathcal{I}' \in \mathfrak{I}(m)\cap \widehat{\mathcal{D}}^{(\ell)}$ that is adjacent to $\mathcal{I}$. Thus, we have $\mathcal{I}\cup \mathcal{I}'\in \mathfrak{I}(m)$, and hence
\begin{eqnarray}\label{event00}
&&~~\frac{1}{|\mathbb{L}_{\ell}^c|}\sum_{i\in \mathbb{L}_{\ell}^c} \mathbb{I}(A_i\in \mathcal{I})\{Y_i-\widehat{q}_{\mathcal{I}}^{(\ell)}(X_i)\}^2+\frac{1}{|\mathbb{L}_{\ell}^c|}\sum_{i\in \mathbb{L}_{\ell}^c} \mathbb{I}(A_i\in \mathcal{I}') \{Y_i-\widehat{q}_{\mathcal{I}'}^{(\ell)}(X_i)\}^2\\ \nonumber
&\le& \frac{1}{|\mathbb{L}_{\ell}^c|}\sum_{i\in \mathbb{L}_{\ell}^c} \mathbb{I}(A_i\in \mathcal{I}\cup \mathcal{I}') \{Y_i-\widehat{q}_{\mathcal{I}\cup\mathcal{I}'}^{(\ell)}(X_i)\}^2-\gamma_n. 
\end{eqnarray}
Notice that the left-hand-side of the above expression is nonnegative. It follows that
\begin{eqnarray*}
	\gamma_n\le \frac{1}{|\mathbb{L}_{\ell}^c|}\sum_{i\in \mathbb{L}_{\ell}^c} \mathbb{I}(A_i\in \mathcal{I}\cup \mathcal{I}') \{Y_i-\widehat{q}_{\mathcal{I}\cup\mathcal{I}'}^{(\ell)}(X_i)\}^2.
\end{eqnarray*}
By definition, we have
\begin{eqnarray*}
\widehat{q}_{\mathcal{I}\cup \mathcal{I}'}^{(\ell)}=\argmin_{q_{\mathcal{I}}\in \mathcal{Q}_{\mathcal{I}}} \frac{1}{n}\sum_{i\in \mathbb{L}_{\ell}^c} \mathbb{I}(A_i\in \mathcal{I}\cup \mathcal{I}') \{Y_i-q_{\mathcal{I}}(X_i)\}^2.
\end{eqnarray*}
It follows that
\begin{eqnarray*}
	\frac{1}{|\mathbb{L}_{\ell}^c|}\sum_{i\in \mathbb{L}_{\ell}^c} \mathbb{I}(A_i\in \mathcal{I}\cup \mathcal{I}') \{Y_i-\widehat{q}_{\mathcal{I}\cup\mathcal{I}'}^{(\ell)}(X_i)\}^2
	\le  \frac{1}{|\mathbb{L}_{\ell}^c|}\sum_{i\in \mathbb{L}_{\ell}^c} \mathbb{I}(A_i\in \mathcal{I}\cup \mathcal{I}') \{Y_i-\widehat{q}_{\mathcal{I}'}^{(\ell)}(X_i)\}^2. 
\end{eqnarray*}
By \eqref{event00}, this further implies that
\begin{eqnarray*}
	\frac{1}{|\mathbb{L}_{\ell}^c|}\sum_{i\in \mathbb{L}_{\ell}^c} \mathbb{I}(A_i\in \mathcal{I})\{Y_i-\widehat{q}_{\mathcal{I}}^{(\ell)}(X_i)\}^2\le \frac{1}{|\mathbb{L}_{\ell}^c|}\sum_{i\in \mathbb{L}_{\ell}^c} \mathbb{I}(A_i\in \mathcal{I})\{Y_i-\widehat{q}_{\mathcal{I}'}^{(\ell)}(X_i)\}^2-\gamma_n,
\end{eqnarray*}
and hence
\begin{eqnarray*}
	\gamma_n\le \frac{1}{|\mathbb{L}_{\ell}^c|}\sum_{i\in \mathbb{L}_{\ell}^c} \mathbb{I}(A_i\in \mathcal{I})\{Y_i-\widehat{q}_{\mathcal{I}'}^{(\ell)}(X_i)\}^2.
\end{eqnarray*}
Under (A2), the function $\widehat{q}_{\mathcal{I}'}$ is uniformly upper bounded from above. 
It thus follows from Cauchy-Schwarz inequality  that
\begin{eqnarray*}
	\gamma_n \le \frac{2}{|\mathbb{L}_{\ell}^c|}\sum_{i\in \mathbb{L}_{\ell}^c} \mathbb{I}(A_i\in \mathcal{I})\{Y_i^2+\widehat{q}_{\mathcal{I}'}^2(X_i)\}\le c_0 n^{-1} \sum_{i\in \mathbb{L}_{\ell}^c} \mathbb{I}(A_i\in \mathcal{I}),
\end{eqnarray*}
for some constant $c_0>0$. 
Using similar arguments in showing \eqref{eqne0}, we can show that with probability at least $1-O(n^{-2})$, the following evens hold for all $\mathcal{I}\in \mathfrak{I}(m)$,
\begin{eqnarray*}
	n^{-1} \sum_{i\in \mathbb{L}_{\ell}^c} \mathbb{I}(A_i\in \mathcal{I})\le c_1 (\sqrt{n^{-1}|\mathcal{I}|\log n}+|\mathcal{I}|),
\end{eqnarray*} 
for some constant $c_1>0$. The right-hand-side shall be larger than or equal to $\gamma_n$. Consequently, we have either $|\mathcal{I}|\ge c_2\gamma_n$ or $|\mathcal{I}|\ge c_2 n\gamma_n^2/\log n$ for some constant $c_2>0$. Under the given condition on $\gamma_n$, we obtain that $|\mathcal{I}|\ge c_2\gamma_n$ for sufficiently large $n$. The proof is hence completed. 
\subsection{Proof of Theorem \ref{thm1}}
Since the number of folds $\mathcal{L}$ is a fixed integer. We will show the assertions in (i) and (ii) holds for each $\ell$, with probability at least $1-O(n^{-2})$. The proof is divided into three parts. In Part 1, we show the consistency of the estimated change point locations and that $|\widehat{\mathcal{D}}^{(\ell)}|\ge |\mathcal{D}_0|$ with probability at least $1-O(n^{-2})$. In Part 2, we prove that $|\widehat{\mathcal{D}}^{(\ell)}|=|\mathcal{D}_0|$ with probability at least $1-O(n^{-2})$ and derive the rate of convergence of the estimated change point locations and the estimated function $Q$. In Part 3, we derive the rate of convergence for the value estimator. 

\smallskip

\noindent \textit{Part 1}. We first show the consistency of the estimated change-point locations. Assume $|\mathcal{D}_0|>1$. Otherwise, the assertion $|\widehat{\mathcal{D}}^{(\ell)}|\ge |\mathcal{D}_0|$ trivially hold. 
Consider the partition $\mathcal{D}=\{[0,1]\}$ which consists of a single interval and a zero function $Q$. By definition, we have
\begin{eqnarray*}
	\sum_{\mathcal{I}\in \widehat{\mathcal{D}}^{(\ell)}} \left( \sum_{i\in \mathbb{L}_{\ell}^c} \mathbb{I}(A_i\in \mathcal{I})\{Y_i-\widehat{q}_{\mathcal{I}}^{(\ell)}(X_i)\}^2 \right)+|\mathbb{L}_{\ell}^c|\gamma_n|\widehat{\mathcal{D}}^{(\ell)}|
	\le \sum_{i\in \mathbb{L}_{\ell}^c} Y_i^2+ |\mathbb{L}_{\ell}^c|\gamma_n.
\end{eqnarray*}
Under the boundedness assumption on $Y$, we obtain that $|\mathbb{L}_{\ell}^c|\gamma_n |\widehat{\mathcal{D}}^{(\ell)}|\le C_0(|\mathbb{L}_{\ell}^c|+\gamma_n)$ for some constant $C_0>0$ and hence
\begin{eqnarray}\label{upperPhat_thm5}
|\widehat{\mathcal{D}}^{(\ell)}|\le  2{C}_0\gamma_n^{-1},
\end{eqnarray}
for sufficiently large $n$, as $\gamma_n\to 0$.

Notice that
\begin{eqnarray*}
	\begin{split}
		\sum_{\mathcal{I}\in \widehat{\mathcal{D}}^{(\ell)}}  \sum_{i\in \mathbb{L}_{\ell}^c} \mathbb{I}(A_i\in \mathcal{I})\{Y_i-\widehat{q}_{\mathcal{I}}^{(\ell)}(X_i)\}^2 \ge \underbrace{ \sum_{\substack{\mathcal{I}\in \widehat{\mathcal{D}}^{(\ell)}} }\sum_{i\in \mathbb{L}_{\ell}^c} \mathbb{I}(A_i\in \mathcal{I})\{Y_i-q_{\mathcal{I},0}(X_i)\}^2}_{\eta_1^*}\\
		+\sum_{\substack{\mathcal{I}\in \widehat{\mathcal{D}}^{(\ell)} }}\sum_{i\in \mathbb{L}_{\ell}^c} \mathbb{I}(A_i\in \mathcal{I}) \{\widehat{q}_{\mathcal{I}}^{(\ell)}(X_i)-q_{\mathcal{I},0}(X_i)\}^2\\
		-2\sum_{\substack{\mathcal{I}\in \widehat{\mathcal{D}}^{(\ell)} }}\left|\sum_{i\in \mathbb{L}_{\ell}^c} \mathbb{I}(A_i\in \mathcal{I})\{Y_i-q_{\mathcal{I},0}(X_i)\}  \{\widehat{q}_{\mathcal{I}}^{(\ell)}(X_i)-q_{\mathcal{I},0}(X_i)\} \right|.
	\end{split}
\end{eqnarray*}
The second line is non-negative. Under Lemmas \ref{lemma1} and \ref{lemma2}, the third line is lower bounded by $-C_1\sum_{\mathcal{I}\in \widehat{\mathcal{D}}^{(\ell)}} (n|\mathcal{I}| )^{p/(p+2\beta)}\log^8 n$ for some constant $C_1>0$ with probability at least $1-O(n^{-2})$. In view of \eqref{upperPhat_thm5}, it can be further lower bounded by $-2C_0C_1 \gamma_n^{-1} n^{p/(p+2\beta)}\log^8 n$. 
By \eqref{upperPhat_thm5} and the given condition on $\gamma_n$, the third line is $o(n)$. It follows that
\begin{eqnarray}\label{eqn:some00}
\sum_{\mathcal{I}\in \widehat{\mathcal{D}}^{(\ell)}}  \sum_{i\in \mathbb{L}_{\ell}^c} \mathbb{I}(A_i\in \mathcal{I})\{Y_i-\widehat{q}_{\mathcal{I}}^{(\ell)}(X_i)\}^2 \ge \eta_1^*+o(n),
\end{eqnarray}
with probability at least $1-O(n^{-2})$. 

Similar to \eqref{eqne0}, we can show that the following events occur with probability at least $1-O(n^{-2})$, 
\begin{eqnarray}\label{eqne2}
	\left|\frac{1}{|\mathbb{L}_{\ell}^c|}\sum_{i\in \mathbb{L}_{\ell}^c} \mathbb{I}(A_i\in\mathcal{I})\{Y_i-Q(X_i,A_i)\}\{Q(X_i,A_i)-q_{\mathcal{I},0}(X_i)\} \right|\\ \nonumber
	\le c_0\left[n^{-1/2}\sqrt{ \Mean \mathbb{I}(A\in \mathcal{I})\{Q(X,A)-q_{\mathcal{I},0}(X)\}^2\log n}+n^{-1}\log n\right],
	\\\label{eqne3}
	\left|\frac{1}{|\mathbb{L}_{\ell}^c|}\sum_{i\in \mathbb{L}_{\ell}^c} \mathbb{I}(A_i\in \mathcal{I})\{Q(X_i,A_i)-q_{\mathcal{I},0}(X_i)\}^2- \Mean \mathbb{I}(A\in \mathcal{I})|Q(X,A)-q_{\mathcal{I}}(X)|^2\right|\\\nonumber\le c_0\left[n^{-1/2}\sqrt{ \Mean \mathbb{I}(A\in \mathcal{I})\{Q(X,A)-q_{\mathcal{I},0}(X)\}^2\log n}+n^{-1}\log n\right],
\end{eqnarray}
for some constant $c_0>0$. For any interval $\mathcal{I}$, the two upper bounds in \eqref{eqne2} and \eqref{eqne3} are $o(1)$. 

It follows that
\begin{eqnarray*}
	\eta_1^*=\sum_{\substack{\mathcal{I}\in \widehat{\mathcal{D}}^{(\ell)} }}\sum_{i\in \mathbb{L}_{\ell}^c} \mathbb{I}(A_i\in \mathcal{I}) \{Y_i-Q(X_i,A_i)\}^2+\sum_{\substack{\mathcal{I}\in \widehat{\mathcal{D}}^{(\ell)} }}\sum_{i\in \mathbb{L}_{\ell}^c} \mathbb{I}(A_i\in \mathcal{I})\{Q(X_i,A_i)-q_{\mathcal{I},0}(X_i)\}^2
	\\+2\sum_{\substack{\mathcal{I}\in \widehat{\mathcal{D}}^{(\ell)} }}\sum_{i\in \mathbb{L}_{\ell}^c} \mathbb{I}(A_i\in \mathcal{I}) \{Y_i-Q(X_i,A_i)\}\{Q(X_i,A_i)-q_{\mathcal{I},0}(X_i)\}\\=\sum_{i\in \mathbb{L}_{\ell}^c} |Y_i-Q(X_i,A_i)|^2+|\mathbb{L}_{\ell}^c|\sum_{\mathcal{I}\in \widehat{\mathcal{D}}^{(\ell)}}\Mean \mathbb{I}(A\in \mathcal{I})|Q(X,A)-q_{\mathcal{I}}(X)|^2+o(n),
\end{eqnarray*}
with probability at least $1-O(n^{-2})$. It follows from \eqref{eqn:some00} that
\begin{eqnarray}\label{eqn:some000}
\begin{split}
\sum_{\mathcal{I}\in \widehat{\mathcal{D}}^{(\ell)}}  \sum_{i\in \mathbb{L}_{\ell}^c} \mathbb{I}(A_i\in \mathcal{I})\{Y_i-\widehat{q}_{\mathcal{I}}^{(\ell)}(X_i)\}^2 \ge \underbrace{\sum_{i\in \mathbb{L}_{\ell}^c} |Y_i-Q(X_i,A_i)|^2}_{\eta_2^*}\\+|\mathbb{L}_{\ell}^c|\sum_{\mathcal{I}\in \widehat{\mathcal{D}}^{(\ell)}}\Mean \mathbb{I}(A\in \mathcal{I})|Q(X,A)-q_{\mathcal{I}}(X)|^2+o(n),
\end{split}	
\end{eqnarray}
with probability at least $1-O(n^{-2})$. 

Let us consider $\eta_2^*$. We observe that
\begin{eqnarray*}
	\eta_2^*=\sum_{\mathcal{I}\in \mathcal{D}_0}\sum_{i\in \mathbb{L}_{\ell}^c} \mathbb{I}(A_i\in \mathcal{I})|Y_i-q_{\mathcal{I},0}(X_i)|^2.
\end{eqnarray*}
By the uniform approximation property of deep neural networks, there exists some $q_{\mathcal{I}}^*\in \mathcal{Q}_{\mathcal{I}}$ such that
\begin{eqnarray*}
	\sum_{i\in \mathbb{L}_{\ell}^c} |q_{\mathcal{I},0}(X_i)-q_{\mathcal{I}}^*(X_i)|^2\propto n(n|\mathcal{I}|)^{-2\beta/(2\beta+p)}.
\end{eqnarray*}
See Part 1 of the proof of Lemma \ref{lemma0} for details. Similar to \eqref{eqne0}, we can show that the following events occur with probability at least $1-O(n^{-2})$, 
\begin{eqnarray*}
	\left|\frac{1}{|\mathbb{L}_{\ell}^c|}\sum_{i\in \mathbb{L}_{\ell}^c} \mathbb{I}(A_i\in\mathcal{I})\{Y_i-q_{\mathcal{I}}(X_i)\}\{q_{\mathcal{I}}(X_i)-q_{\mathcal{I}}^*(X_i)\} \right|\le \frac{c_0\sqrt{|\mathcal{I}|\log n}}{\sqrt{n}}(n|\mathcal{I}|)^{-\beta/(2\beta+p)},
\end{eqnarray*}
for some constant $c_0>0$ and any $\mathcal{I}\in \mathcal{D}_0$. It follows that
\begin{eqnarray*}
	\eta_2^*-\sum_{\mathcal{I}\in \mathcal{D}_0}\sum_{i\in \mathbb{L}_{\ell}^c} \mathbb{I}(A_i\in \mathcal{I})|Y_i-q_{\mathcal{I}}^*(X_i)|^2\ge -\sum_{\mathcal{I}\in \mathcal{D}_0}\sum_{i\in \mathbb{L}_{\ell}^c} \mathbb{I}(A_i\in \mathcal{I})|q_{\mathcal{I},0}(X_i)-q_{\mathcal{I}}^*(X_i)|^2\\
	-2\sum_{\mathcal{I}\in \mathcal{D}_0}\left|\sum_{i\in \mathbb{L}_{\ell}^c} \mathbb{I}(A_i\in\mathcal{I})\{Y_i-q_{\mathcal{I}}(X_i)\}\{q_{\mathcal{I}}(X_i)-q_{\mathcal{I}}^*(X_i)\} \right|\ge -\bar{c} n^{p/(2\beta+p)},
\end{eqnarray*}
for some constant $\bar{c}>0$. This together with \eqref{eqn:some000} yields that
\begin{eqnarray}\label{eqne1}
\begin{split}
	\sum_{\mathcal{I}\in \widehat{\mathcal{D}}^{(\ell)}}  \sum_{i\in \mathbb{L}_{\ell}^c} \mathbb{I}(A_i\in \mathcal{I})\{Y_i-\widehat{q}_{\mathcal{I}}^{(\ell)}(X_i)\}^2 \ge \sum_{\mathcal{I}\in \mathcal{D}_0}\sum_{i\in \mathbb{L}_{\ell}^c} \mathbb{I}(A_i\in \mathcal{I})|Y_i-q_{\mathcal{I}}^*(X_i)|^2\\+|\mathbb{L}_{\ell}^c|\sum_{\mathcal{I}\in \widehat{\mathcal{D}}^{(\ell)}}\Mean \mathbb{I}(A\in \mathcal{I})|Q(X,A)-q_{\mathcal{I},0}(X)|^2+o(n)+O\{n^{p/(2\beta+p)}\},
\end{split}	
\end{eqnarray}
with probability at least $1-O(n^{-2})$. 

Let $K=|\mathcal{D}_0|$. For any integer $k$ such that $1\le k\le K-1$, let $\tau_{0,k}^*$ be the change point location that satisfies $\tau_{0,k}^*=i/m$ for some integer $i$ and that $|\tau_{0,k}-\tau_{0,k}^*|<m^{-1}$. Denoted by $\mathcal{D}^*$ the oracle partition formed by the change point locations $\{\tau_{0,k}^*\}_{k=1}^{K-1}$. Set $\tau_{0,0}^*=0$, $\tau_{0,K}^*=1$ and $q_{[\tau_{0,k-1}^*,\tau_{0,k}^*)}^{**}=q_{[\tau_{0,k-1},\tau_{0,k})}^*$ for $1\le k\le K-1$. Let $\Delta_k=[\tau_{0,k-1}^*,\tau_{0,k}^*) \cap [\tau_{0,k-1},\tau_{0,k})^c$ for $1\le k\le K-1$ and $\Delta_{K}=[\tau_{0,K-1}^*,1] \cap [\tau_{0,K-1},1]^c$. The length of each interval $\Delta_k$ is at most $m^{-1}$. 
It follows that
\begin{eqnarray*}
&&\left( \sum_{\mathcal{I}\in \mathcal{D}^*}\left[\sum_{i\in \mathbb{L}_{\ell}^c} \mathbb{I}(A_i\in \mathcal{I})\{Y_i-q_{\mathcal{I}}^{**}(X_i)\}^2 \right]+\gamma_n |\mathbb{L}_{\ell}^c| |\mathcal{D}^*| \right)\\ \nonumber
&-&\left( \sum_{\mathcal{I}\in \mathcal{D}_0}\left[ \sum_{i\in \mathbb{L}_{\ell}^c} \mathbb{I}(A_i\in \mathcal{I})\{Y_i-q_{\mathcal{I}}^*(X_i)\}^2\right]+\gamma_n |\mathbb{L}_{\ell}^c||\mathcal{D}_0| \right)\\ \nonumber
\le&& 2\sum_{k=1}^K \sum_{i\in \mathbb{L}_{\ell}^c} \mathbb{I}(A_i\in \Delta_k) \left\{Y_i^2+\sup_{\mathcal{I}\subseteq [0,1]} q_{\mathcal{I}}^{*2}(X_i) \right\}.
\end{eqnarray*} 
Since $Y$ is a bounded variable, $q_{\mathcal{I}}^*$ is uniformly bounded for any $\mathcal{I}$. The right-hand-side is upper bounded by $\sum_{k=1}^K \sum_{i\in \mathbb{L}_{\ell}^c} \mathbb{I}(A_i\in \Delta_k)$. Similar to \eqref{eqne0}, The later is upper bounded by $O(\log n)$, with probability at least $1-O(n^{-2})$.It follows that
\begin{eqnarray}\label{lowerboundeq2}
&&\left( \sum_{\mathcal{I}\in \mathcal{D}^*}\left[ \sum_{i\in \mathbb{L}_{\ell}^c} \mathbb{I}(A_i\in \mathcal{I})\{Y_i-q_{\mathcal{I}}^{**}(X_i)\}^2 \right]+\gamma_n |\mathbb{L}_{\ell}^c| |\mathcal{D}^*| \right)\\ \nonumber
&-&\left( \sum_{\mathcal{I}\in \mathcal{D}_0}\left[ \sum_{i\in \mathbb{L}_{\ell}^c} \mathbb{I}(A_i\in \mathcal{I})\{Y_i-q_{\mathcal{I}}^*(X_i)\}^2\right]+\gamma_n|\mathbb{L}_{\ell}^c||\mathcal{D}_0|\right)\le O(\log n),
\end{eqnarray}
with probability at least $1-O(n^{-2})$. By definition,
\begin{eqnarray}\label{upperbound}
\begin{split}
	\sum_{\mathcal{I}\in \widehat{\mathcal{D}}^{(\ell)}} \sum_{i\in \mathbb{L}_{\ell}^c} \mathbb{I}(A_i\in \mathcal{I})\{Y_i-\widehat{q}_{\mathcal{I}}^{(\ell)}(X_i)\}^2+\gamma_n|\mathbb{L}_{\ell}^c||\widehat{\mathcal{D}}^{(\ell)}|\\\le
	\sum_{\mathcal{I}\in \mathcal{D}^*}\sum_{i\in \mathbb{L}_{\ell}^c} \mathbb{I}(A_i\in \mathcal{I})\{Y_i-q_{\mathcal{I}}^{**}(X_i)\}^2+\gamma_n|\mathbb{L}_{\ell}^c| |\mathcal{D}^*|.
\end{split}	
\end{eqnarray}
Combining this together with \eqref{lowerboundeq2} yields that
\begin{eqnarray*}
	\sum_{\mathcal{I}\in \widehat{\mathcal{D}}^{(\ell)}} \sum_{i\in \mathbb{L}_{\ell}^c} \mathbb{I}(A_i\in \mathcal{I})\{Y_i-\widehat{q}_{\mathcal{I}}^{(\ell)}(X_i)\}^2+\gamma_n|\mathbb{L}_{\ell}^c||\widehat{\mathcal{D}}^{(\ell)}|\\\le
	\sum_{\mathcal{I}\in \mathcal{D}_0}\sum_{i\in \mathbb{L}_{\ell}^c} \mathbb{I}(A_i\in \mathcal{I})\{Y_i-q_{\mathcal{I}}^{*}(X_i)\}^2+\gamma_n|\mathbb{L}_{\ell}^c| |\mathcal{D}_0|+O(\log n).
\end{eqnarray*}
It follows from \eqref{eqne1} and the condition $\gamma_n\to0$ that
\begin{eqnarray}\label{eqn:event1}
	\sum_{\mathcal{I}\in \widehat{\mathcal{D}}^{(\ell)}}\Mean \mathbb{I}(A\in \mathcal{I})|Q(X,A)-q_{\mathcal{I},0}(X)|^2=o(1),
\end{eqnarray}
with probability at least $1-O(n^{-2})$. Under the event defined above, we show that $\max_{\tau\in J(\mathcal{D}_0)} \min_{\hat{\tau}\in J(\widehat{\mathcal{D}}^{(\ell)})} |\hat{\tau}-\tau|\le \delta$ for any constant $\delta>0$. This yields the consistency of our estimated change point locations. 

Specifically, under the condition that $q_{\mathcal{I}_1,0}\neq q_{\mathcal{I}_2,0}$ for any adjacent $\mathcal{I}_1,\mathcal{I}_2\in \mathcal{D}_0$, we have $\Mean |q_{\mathcal{I}_1,0}(X)- q_{\mathcal{I}_2,0}(X)|^2>0$. Let $\delta_0$ denote the minimum distance between two change point locations. Since the change points are fixed, $\delta_0$ is a fixed positive value. For a given $0<\delta<\delta_0$, suppose $\max_{\tau\in J(\mathcal{D}_0)} \min_{\hat{\tau}\in J(\widehat{\mathcal{D}}^{(\ell)})} |\hat{\tau}-\tau|> \delta$. Then there exists a change point $\tau_0$ and $\mathcal{I}\in \widehat{\mathcal{D}}^{(\ell)}$ such that $\tau_0\in \mathcal{I}$, $|\mathcal{I}|\ge 2\delta$ and that $\min(|a-\tau_0|,|b-\tau_0|)\ge \delta$ where $a,b$ correspond to the endpoints of the interval $\mathcal{I}$. Under the event defined in \eqref{eqn:event1}, we have
\begin{eqnarray}\label{eqn:event3}
	\Mean \mathbb{I}(A\in [a,b])|Q(X,A)-q_{\mathcal{I},0}(X)|^2=o(1).
\end{eqnarray}
Since $\delta_0>\delta$, the conditional mean function $Q$ is a piecewise function of $A$ in the intervals $[a,\tau_0]$ and $[\tau_0,b]$. The left-hand-side thus equals
\begin{eqnarray*}
	\Mean \mathbb{I}(A\in [\tau_0,b])|q_{[\tau_0,b],0}(X)-q_{\mathcal{I},0}(X)|^2+\Mean \mathbb{I}(A\in [a,\tau_0])|q_{[a,\tau_0],0}(X)-q_{\mathcal{I},0}(X)|^2.
\end{eqnarray*}
The function $q_{\mathcal{I},0}$ that minimizes the above objective is given by
\begin{eqnarray*}
	\{\Mean \mathbb{I}(A\in [a,b]|X)\}^{-1} [q_{[a,\tau_0],0}(X) \Mean \{\mathbb{I}(A\in [a,\tau_0])|X\} +q_{[\tau_0,b],0}(X)\Mean \{\mathbb{I}(A\in [\tau_0,b])|X\} ].
\end{eqnarray*}
Consequently, the left-hand-side of \eqref{eqn:event3} is greater than or equal to
\begin{eqnarray*}
	\Mean \{\mathbb{I}(A\in [\tau_0,b])|X\} \{\mathbb{I}(A\in [a,\tau_0])|X\} |q_{[\tau_0,b],0}(X)-q_{[a,\tau_0],0}(X)|^2,
\end{eqnarray*}
which is not to decay to zero since $\min(|a-\tau_0|,|b-\tau_0|)\ge \delta$ and that $q_{\mathcal{I}_1,0}\neq q_{\mathcal{I}_2,0}$ for any adjacent $\mathcal{I}_1,\mathcal{I}_2\in \mathcal{D}_0$. This contradicts \eqref{eqn:event3}. As such, we obtain that $\max_{\tau\in J(\mathcal{D}_0)} \min_{\hat{\tau}\in J(\widehat{\mathcal{D}}^{(\ell)})} |\hat{\tau}-\tau|\le \delta$ for any sufficiently small $\delta$. This yields the consistency of the estimated change point locations. It also implies that $|\widehat{\mathcal{D}}^{(\ell)}|\ge |\mathcal{D}_0|$ with probability at least $1-O(n^{-2})$. This completes the proof of Part 1. 

\smallskip

\noindent \textit{Part 2}. In this part, we show $|\widehat{\mathcal{D}}^{(\ell)}|=|\mathcal{D}_0|$ with probability at least $1-O(n^{-2})$ and derive the rate of convergence of the estimated change point locations. Similar to \eqref{eqne1} and \eqref{lowerboundeq2}, with a more refined analysis (see Part 1 of the proof), we obtain that 
\begin{eqnarray*}
	\begin{split}
		\sum_{\mathcal{I}\in \widehat{\mathcal{D}}^{(\ell)}}  \sum_{i\in \mathbb{L}_{\ell}^c} \mathbb{I}(A_i\in \mathcal{I})\{Y_i-\widehat{q}_{\mathcal{I}}(X_i)\}^2 \ge \sum_{\mathcal{I}\in \mathcal{D}^*}\sum_{i\in \mathbb{L}_{\ell}^c} \mathbb{I}(A_i\in \mathcal{I})|Y_i-q_{\mathcal{I}}^{**}(X_i)|^2\\+|\mathbb{L}_{\ell}^c|\sum_{\mathcal{I}\in \widehat{\mathcal{D}}^{(\ell)}}\Mean \mathbb{I}(A\in \mathcal{I})|Q(X,A)-q_{\mathcal{I},0}(X)|^2-C_1|\widehat{\mathcal{D}}^{(\ell)}|^{\beta/(2p+\beta)}n^{p/(p+2\beta)}\log^8 n+O(n^{p/(2\beta+p)})\\
		-2c_0 |\mathbb{L}_{\ell}^c|^{1/2} \sum_{\mathcal{I} \in \widehat{\mathcal{D}}^{(\ell)}}\sqrt{ \Mean \mathbb{I}(A\in \mathcal{I})\{Q(X,A)-q_{\mathcal{I},0}(X)\}^2\log n}-2c_0 |\widehat{\mathcal{D}}^{(\ell)}| \log n.
	\end{split}	
\end{eqnarray*}
with probability at least $1-O(n^{-2})$. By Cauchy-Schwarz inequality, the third line is lower bounded by
\begin{eqnarray*}
	-\frac{|\mathbb{L}_{\ell}^c|}{2}\sum_{\mathcal{I}\in \widehat{\mathcal{D}}^{(\ell)}}\Mean \mathbb{I}(A\in \mathcal{I})|Q(X,A)-q_{\mathcal{I},0}(X)|^2-2(c_0+c_0^2)|\widehat{\mathcal{D}}^{(\ell)}| \log n.
\end{eqnarray*}
It follows that
\begin{eqnarray*}
	\begin{split}
		\sum_{\mathcal{I}\in \widehat{\mathcal{D}}^{(\ell)}}  \sum_{i\in \mathbb{L}_{\ell}^c} \mathbb{I}(A_i\in \mathcal{I})\{Y_i-\widehat{q}_{\mathcal{I}}(X_i)\}^2 \ge \sum_{\mathcal{I}\in \mathcal{D}^*}\sum_{i\in \mathbb{L}_{\ell}^c} \mathbb{I}(A_i\in \mathcal{I})|Y_i-q_{\mathcal{I}}^{**}(X_i)|^2\\+\frac{|\mathbb{L}_{\ell}^c|}{2}\sum_{\mathcal{I}\in \widehat{\mathcal{D}}^{(\ell)}}\Mean \mathbb{I}(A\in \mathcal{I})|Q(X,A)-q_{\mathcal{I},0}(X)|^2-C_1|\widehat{\mathcal{D}}^{(\ell)}|^{\beta/(2p+\beta)}n^{p/(p+2\beta)}\log^8 n\\ -2(c_0+c_0^2)|\widehat{\mathcal{D}}^{(\ell)}|\log n
		+O(n^{p/(2\beta+p)}).
	\end{split}	
\end{eqnarray*}
 This together with \eqref{upperbound} yields that
\begin{eqnarray*}
	\frac{|\mathbb{L}_{\ell}^c|}{2}\sum_{\mathcal{I}\in \widehat{\mathcal{D}}^{(\ell)}}\Mean \mathbb{I}(A\in \mathcal{I})|Q(X,A)-q_{\mathcal{I},0}(X)|^2\le C_1|\widehat{\mathcal{D}}^{(\ell)}|^{\beta/(2p+\beta)}n^{p/(p+2\beta)}\log^8 n\\+O(n^{p/(2\beta+p)})+n\gamma_n (|\mathcal{D}_0|-|\widehat{\mathcal{D}}^{(\ell)}|)+2(c_0+c_0^2) |\widehat{\mathcal{D}}^{(\ell)}| \log n.
\end{eqnarray*}
Under the given condition on $\gamma_n$, we obtain that $|\mathcal{\widehat{D}}^{(\ell)}|\le |\mathcal{D}_0|$. Combining this together with $|\mathcal{\widehat{D}}^{(\ell)}|\ge |\mathcal{D}_0|$, we obtain that $|\mathcal{\widehat{D}}^{(\ell)}|=|\mathcal{D}_0|$. As such, we have that
\begin{eqnarray*}
	\sum_{\mathcal{I}\in \widehat{\mathcal{D}}^{(\ell)}}\Mean \mathbb{I}(A\in \mathcal{I})|Q(X,A)-q_{\mathcal{I},0}(X)|^2\propto n^{-2\beta/(p+2\beta)}\log^8 n
\end{eqnarray*}
Using similar arguments in establishing the consistency of the estimated change point locations, we can show that under the above event, we have that $\max_{\tau\in J(\mathcal{D}_0)} \min_{\hat{\tau}\in J(\widehat{\mathcal{D}}^{(\ell)})} |\hat{\tau}-\tau|\propto n^{-2\beta/(p+2\beta)}\log^8 n$. This completes the proof of this part. 

\smallskip

\noindent \textit{Part 3}. 
For any target policy $\pi$, we  define a random policy $\pi_{\widehat{\mathcal{D}}^{(\ell)}}$ according to the partition $\widehat{\mathcal{D}}^{(\ell)}$ as follows:
\begin{eqnarray*}
	\pi_{\widehat{\mathcal{D}}^{(\ell)}}(a|x)=\sum_{\mathcal{I}\subseteq \widehat{\mathcal{D}}^{(\ell)}} \mathbb{I}\{\pi(x)\in \mathcal{I},a\in \mathcal{I}\} \frac{b(a|x)}{b(\mathcal{I}|x)},
\end{eqnarray*}
where $b(\mathcal{I}|x)$ denotes the propensity score function $\prob(A\in \mathcal{I}|X=x)$. 
Note that $\int_0^1 \pi_{\widehat{\mathcal{D}}^{(\ell)}}(a|x) da=\sum_{\mathcal{I}\subseteq \widehat{\mathcal{D}}^{(\ell)}} \mathbb{I}\{\pi(x)\in \mathcal{I}\}=1$ for any $x$. Consequently, $\pi_{\widehat{\mathcal{D}}^{(\ell)}}$ is a valid random policy. 

Since the behavior policy is known, the proposed doubly-robust estimator corresponds to an unbiased estimator for $\mathcal{L}^{-1} \sum_{\ell=1}^{\mathcal{L}} V(\pi_{\widehat{\mathcal{D}}^{(\ell)}})$. Using similar arguments in the causal inference literature on deriving the asymptotic property of doubly-robust estimators \citep{chernozhukov2017double}, we can show that
\begin{eqnarray*}
	\widehat{V}(\pi)-\frac{1}{\mathcal{L}} \sum_{\ell=1}^{\mathcal{L}} V(\pi_{\widehat{\mathcal{D}}^{(\ell)}})=O_p(n^{-1/2}).
\end{eqnarray*}
It suffices to show $\mathcal{L}^{-1} \sum_{\ell=1}^{\mathcal{L}} \{V(\pi_{\widehat{\mathcal{D}}^{(\ell)}})-V(\pi)\}=O_p\{n^{-2\beta/(2\beta+p)}\log^8 n\}$, or equivalently, $V(\pi_{\widehat{\mathcal{D}}^{(\ell)}})-V(\pi)=O_p\{n^{-2\beta/(2\beta+p)}\log^8 n\}$. 

Based on the results obtained in the first two parts, it follows from Cauchy-Schwarz inequality that 
\begin{eqnarray}\label{eqn:uniconv}
\begin{split}
\sum_{\mathcal{I}\in \widehat{\mathcal{D}}^{(\ell)}}\Mean \left[\mathbb{I}(A\in \mathcal{I})|Q(X,A)-\widehat{q}_{\mathcal{I}}^{(\ell)}(X)|^2|X\right]\le 2\sum_{\mathcal{I}\in \widehat{\mathcal{D}}^{(\ell)}}\Mean \mathbb{I}(A\in \mathcal{I})|Q(X,A)-q_{\mathcal{I},0}(X)|^2\\+2\sum_{\mathcal{I}\in \widehat{\mathcal{D}}^{(\ell)}}\Mean \left[\mathbb{I}(A\in \mathcal{I})|\widehat{q}_{\mathcal{I}}^{(\ell)}(X)-q_{\mathcal{I},0}(X)|^2|X\right]\propto n^{-2\beta/(p+2\beta)}\log^8 n.
\end{split}	
\end{eqnarray} 
Note that
\begin{eqnarray*}
	V(\pi_{\widehat{\mathcal{D}}^{(\ell)}})=\Mean \int_{[0,1]} Q(X,a) \sum_{\mathcal{I}\subseteq \widehat{\mathcal{D}}^{(\ell)}} \mathbb{I}\{\pi(X)\in \mathcal{I},a\in \mathcal{I}\} \frac{b(a|X)}{b(\mathcal{I}|X)}da\\=\sum_{\mathcal{I}_0\in \mathcal{D}_0} \Mean q_{\mathcal{I}_0}(X)\sum_{\mathcal{I}\subseteq \widehat{\mathcal{D}}^{(\ell)}} \mathbb{I}\{\pi(X)\in \mathcal{I}\}\frac{b(\mathcal{I}\cap \mathcal{I}_0|X)}{b(\mathcal{I}|X)}. 
\end{eqnarray*}
Similarly, we can show
\begin{eqnarray*}
V(\pi)=\sum_{\mathcal{I}_0\in \mathcal{D}_0} \Mean q_{\mathcal{I}_0}(X) \mathbb{I}\{\pi(X)\in \mathcal{I}_0\}.
\end{eqnarray*}
It follows that
\begin{eqnarray*}
	|V(\pi_{\widehat{\mathcal{D}}^{(\ell)}})-V(\pi)|\le \sum_{\mathcal{I}_0\in \mathcal{D}_0} \Mean |q_{\mathcal{I}_0}(X)| \left| \mathbb{I}\{\pi(X)\in \mathcal{I}_0\}- \sum_{\mathcal{I}\subseteq \widehat{\mathcal{D}}^{(\ell)}} \mathbb{I}\{\pi(X)\in \mathcal{I}\}\frac{b(\mathcal{I}\cap \mathcal{I}_0|X)}{b(\mathcal{I}|X)}\right|.
\end{eqnarray*}
As $q_{\mathcal{I}_0}$ is uniformly bounded, the left-hand-side is upper bounded by
\begin{eqnarray}\label{eqn:bound}
	\sum_{\mathcal{I}_0\in \mathcal{D}_0} \Mean \left| \mathbb{I}\{\pi(X)\in \mathcal{I}_0\}- \sum_{\mathcal{I}\subseteq \widehat{\mathcal{D}}^{(\ell)}} \mathbb{I}\{\pi(X)\in \mathcal{I}\}\frac{b(\mathcal{I}\cap \mathcal{I}_0|X)}{b(\mathcal{I}|X)}\right|.
\end{eqnarray}
Based on the results obtained in Part 2, for each $\mathcal{I}_0\in \mathcal{D}_0$, there exists some $\mathcal{I}_0^{(\ell)}$ where the Lebesgue measure of the difference $\mathcal{I}_0\cap (\mathcal{I}_0^{(\ell)})^c + \mathcal{I}_0^c \cap \mathcal{I}_0^{(\ell)}$ is upper bounded by $O\{n^{-2\beta/(2\beta+p)}\log^8 n\}$, with probability at least $1-O(n^{-2})$. The upper bound in \eqref{eqn:bound} is $O\{n^{-2\beta/(2\beta+p)}\log^8 n\}$, under the positivity assumption and the assumption that $\prob(\pi(X)\in [\tau_0-\epsilon,\tau_0+\epsilon])=O(\epsilon)$ for any $\tau_0\in J(\mathcal{D}_0)$ and sufficiently small $\epsilon>0$. This completes the proof. 

\subsection{Proof of Theorem \ref{thm2}}
We break the proof into two parts. In Part 1, we introduce an auxiliary lemma and present its proof. In Part 2, we derive the convergence rate of the proposed value estimator. 

\smallskip

\noindent \textit{Part 1.} We first introduce the following lemma.
\begin{lemma}\label{lemmacontinuous}
	For any interval $\mathcal{I}\in \mathfrak{I}(m)$ with $|\mathcal{I}|\gg \gamma_n$ and any interval $\mathcal{I}'\in \widehat{\mathcal{D}}^{(\ell)}$ with $\mathcal{I}\subseteq \mathcal{I}'$, we have with probability approaching $1$ that
	\begin{eqnarray*}
		\Mean |q_{\mathcal{I},0}(X)-q_{\mathcal{I}',0}(X)|^2\le \bar{C} |\mathcal{I}|^{-1}\gamma_n,
	\end{eqnarray*}
	for some constant $\bar{C}>0$. 
\end{lemma}	
We next prove Lemma \ref{lemmacontinuous}. 
For a given interval $\mathcal{I}'\in \widehat{\mathcal{D}}^{(\ell)}$, the set of intervals $\mathcal{I}$ considered in Lemma \ref{lemmacontinuous} can be classified into the following three categories. 

\noindent \textit{Category 1:} $\mathcal{I}=\mathcal{I}'$. It is immediate to see that $q_{\mathcal{I}}=q_{\mathcal{I}'}$ and the assertion automatically holds. 

\noindent \textit{Category 2:} There exists another interval $\mathcal{I}^*\in \mathfrak{I}(m)$ that satisfies $\mathcal{I}'=\mathcal{I}^*\cup \mathcal{I}$. Notice that the partition $\widehat{\mathcal{D}}^{(\ell)*}=\widehat{\mathcal{D}}^{(\ell)} \cup \{\mathcal{I}^* \} \cup \mathcal{I}-\{\mathcal{I}' \}$ forms another partition. By definition, we have
\begin{eqnarray*}
	&&\frac{1}{|\mathbb{L}_{\ell}^c|}\sum_{i\in \mathbb{L}_{\ell}^c}\sum_{\mathcal{I}_0\in \widehat{\mathcal{D}}^{(\ell)*}} \mathbb{I}(A_i\in \mathcal{I}_0) \{Y_i-\widehat{q}_{\mathcal{I}_0}(X_i)\}^2+\gamma_n |\widehat{\mathcal{D}}^{(\ell)*}|\\
	\ge && \frac{1}{|\mathbb{L}_{\ell}^c|}\sum_{i\in \mathbb{L}_{\ell}^c}\sum_{\mathcal{I}_0\in \widehat{\mathcal{D}}^{(\ell)}} \mathbb{I}(A_i\in \mathcal{I}_0) \{Y_i-\widehat{q}_{\mathcal{I}_0}(X_i)\}^2+\gamma_n |\widehat{\mathcal{D}}^{(\ell)}|,
\end{eqnarray*}
and hence
\begin{eqnarray*}
	\frac{1}{|\mathbb{L}_{\ell}^c|}\sum_{i\in \mathbb{L}_{\ell}^c} \mathbb{I}(A_i\in \mathcal{I}) \{Y_i-\widehat{q}_{\mathcal{I}}(X_i)\}^2+\frac{1}{|\mathbb{L}_{\ell}^c|}\sum_{i\in \mathbb{L}_{\ell}^c} \mathbb{I}(A_i\in \mathcal{I}^*) \{Y_i-\widehat{q}_{\mathcal{I}^*}(X_i)\}^2\\ 
	\ge \frac{1}{|\mathbb{L}_{\ell}^c|}\sum_{i\in \mathbb{L}_{\ell}^c} \mathbb{I}(A_i\in \mathcal{I}') \{Y_i-\widehat{q}_{\mathcal{I}'}(X_i)\}^2-\gamma_n.
\end{eqnarray*}
It follows from the definition of $\widehat{q}_{\mathcal{I}^*}$ that
\begin{eqnarray*}
	\frac{1}{|\mathbb{L}_{\ell}^c|}\sum_{i\in \mathbb{L}_{\ell}^c} \mathbb{I}(A_i\in \mathcal{I}^*) \{Y_i-\widehat{q}_{\mathcal{I}^*}(X_i)\}^2\le \frac{1}{|\mathbb{L}_{\ell}^c|}\sum_{i\in \mathbb{L}_{\ell}^c} \mathbb{I}(A_i\in \mathcal{I}^*) \{Y_i-\widehat{q}_{\mathcal{I}'}(X_i)\}^2.
\end{eqnarray*}
Therefore, we obtain
\begin{eqnarray}\label{cat2eq1}
\frac{1}{|\mathbb{L}_{\ell}^c|}\sum_{i\in \mathbb{L}_{\ell}^c} \mathbb{I}(A_i\in \mathcal{I}) \{Y_i-\widehat{q}_{\mathcal{I}}(X_i)\}^2 \ge
\frac{1}{|\mathbb{L}_{\ell}^c|}\sum_{i\in \mathbb{L}_{\ell}^c} \mathbb{I}(A_i\in \mathcal{I}) \{Y_i-\widehat{q}_{\mathcal{I}'}(X_i)\}^2-\gamma_n.
\end{eqnarray}

\noindent \textit{Category 3: }There exist two intervals $\mathcal{I}^*,\mathcal{I}^{**}\in \mathfrak{I}(m)$ that satisfy $\mathcal{I}'=\mathcal{I}^*\cup \mathcal{I} \cup \mathcal{I}^{**}$. Using similar arguments in proving \eqref{cat2eq1}, we can show that
\begin{eqnarray*}
	\frac{1}{|\mathbb{L}_{\ell}^c|}\sum_{i\in \mathbb{L}_{\ell}^c} \mathbb{I}(A_i\in \mathcal{I}) \{Y_i-\widehat{q}_{\mathcal{I}}(X_i)\}^2 \ge 
	\frac{1}{|\mathbb{L}_{\ell}^c|}\sum_{i\in \mathbb{L}_{\ell}^c} \mathbb{I}(A_i\in \mathcal{I}) \{Y_i-\widehat{q}_{\mathcal{I}'}(X_i)\}^2-2\gamma_n.
\end{eqnarray*}
Hence, regardless of whether $\mathcal{I}$ belongs to Category 2, or it belongs to Category 3, we have
\begin{eqnarray}\label{cat2eq2}
\frac{1}{|\mathbb{L}_{\ell}^c|}\sum_{i\in \mathbb{L}_{\ell}^c} \mathbb{I}(A_i\in \mathcal{I}) \{Y_i-\widehat{q}_{\mathcal{I}}(X_i)\}^2 \ge \frac{1}{|\mathbb{L}_{\ell}^c|}\sum_{i\in \mathbb{L}_{\ell}^c} \mathbb{I}(A_i\in \mathcal{I}) \{Y_i-\widehat{q}_{\mathcal{I}'}(X_i)\}^2-2\gamma_n.
\end{eqnarray}
Notice that for any interval $\mathcal{I}_0$,
\begin{eqnarray*}
	\frac{1}{|\mathbb{L}_{\ell}^c|}\sum_{i\in \mathbb{L}_{\ell}^c} \mathbb{I}(A_i\in \mathcal{I}_0) \{Y_i-\widehat{q}_{\mathcal{I}_0}(X_i)\}^2-\Mean [\mathbb{I}(A\in \mathcal{I}_0) \{Y-\widehat{q}_{\mathcal{I}_0}(X)\}^2|\{O_i\}_{i\in \mathbb{L}_{\ell}^c}]\\
	=\frac{1}{|\mathbb{L}_{\ell}^c|}\sum_{i\in \mathbb{L}_{\ell}^c} \mathbb{I}(A_i\in \mathcal{I}_0) \{\widehat{q}_{\mathcal{I}_0}(X_i)-q_{\mathcal{I}_0,0}(X_i)\}\{q_{\mathcal{I},0}(X_i)-\widehat{q}_{\mathcal{I}_0,0}(X_i)\}^2\\+\frac{1}{|\mathbb{L}_{\ell}^c|}\sum_{i\in \mathbb{L}_{\ell}^c} \mathbb{I}(A_i\in \mathcal{I}_0) \{Y_i-\widehat{q}_{\mathcal{I}_0}(X_i)\}^2-\Mean [\mathbb{I}(A\in \mathcal{I}_0) \{\widehat{q}_{\mathcal{I}_0}(X_i)-\widehat{q}_{\mathcal{I}_0}(X)\}^2|\{O_i\}_{i\in \mathbb{L}_{\ell}^c}].
\end{eqnarray*}
Using similar arguments in bounding the stochastic error term in Part 2 of the proof of Lemma \ref{lemma0}, we can show with probability approaching $1$ that the right-hand-side is of the order $O\{n^{-2\beta/(2\beta+p)}\log^8 n\}$, for any $\mathcal{I}_0\in \mathfrak{I}(m)$. As such, we obtain with probability approaching $1$ that
\begin{eqnarray*}
	\frac{1}{|\mathbb{L}_{\ell}^c|}\sum_{i\in \mathbb{L}_{\ell}^c} \mathbb{I}(A_i\in \mathcal{I}) \{Y_i-\widehat{q}_{\mathcal{I}}(X_i)\}^2=\Mean [\mathbb{I}(A\in \mathcal{I}) \{Y-\widehat{q}_{\mathcal{I}}(X)\}^2|\{O_i\}_{i\in \mathbb{L}_{\ell}^c}]\\
	+O(1)|\mathcal{I}|(n|\mathcal{I}|)^{-2\beta/(2\beta+p)}\log^8 n,\\
	\frac{1}{|\mathbb{L}_{\ell}^c|}\sum_{i\in \mathbb{L}_{\ell}^c} \mathbb{I}(A_i\in \mathcal{I}) \{Y_i-\widehat{q}_{\mathcal{I}'}(X_i)\}^2=\Mean [\mathbb{I}(A\in \mathcal{I}) \{Y-\widehat{q}_{\mathcal{I}'}(X)\}^2|\{O_i\}_{i\in \mathbb{L}_{\ell}^c}]\\
	+O(1)|\mathcal{I}|(n|\mathcal{I}|)^{-2\beta/(2\beta+p)}\log^8 n,
\end{eqnarray*}
where $O(1)$ denotes some universal positive constant. Combining these together with \eqref{cat2eq2} yields
\begin{eqnarray*}
	\Mean [\mathbb{I}(A\in \mathcal{I}) \{Y-\widehat{q}_{\mathcal{I}}(X)\}^2|\{O_i\}_{i\in \mathbb{L}_{\ell}^c}]\ge \Mean [\mathbb{I}(A\in \mathcal{I}) \{Y-\widehat{q}_{\mathcal{I}'}(X)\}^2|\{O_i\}_{i\in \mathbb{L}_{\ell}^c}]\\-2\gamma_n+O(1)|\mathcal{I}|(n|\mathcal{I}|)^{-2\beta/(2\beta+p)}\log^8 n,
\end{eqnarray*}
for any $\mathcal{I}$ and $\mathcal{I}'$, with probability approaching $1$. Note that $q_{\mathcal{I},0}$ satisfies $\Mean [\mathbb{I}(A\in \mathcal{I})\{Y-q_{\mathcal{I},0}(X)\}|X]=0$. We have
\begin{eqnarray*}
	\Mean [\mathbb{I}(A\in \mathcal{I}) \{q_{\mathcal{I},0}(X)-\widehat{q}_{\mathcal{I}}(X)\}^2|\{O_i\}_{i\in \mathbb{L}_{\ell}^c}]\ge \Mean [\mathbb{I}(A\in \mathcal{I}) \{q_{\mathcal{I},0}(X)-\widehat{q}_{\mathcal{I}'}(X)\}^2|\{O_i\}_{i\in \mathbb{L}_{\ell}^c}]\\-2\gamma_n+O(1)|\mathcal{I}|(n|\mathcal{I}|)^{-2\beta/(2\beta+p)}\log^8 n.
\end{eqnarray*}
Consider the first term on the right-hand-side. Note that
\begin{eqnarray*}
	\Mean [\mathbb{I}(A\in \mathcal{I}) \{q_{\mathcal{I},0}(X)-\widehat{q}_{\mathcal{I}'}(X)\}^2|\{O_i\}_{i\in \mathbb{L}_{\ell}^c}]=\Mean [\mathbb{I}(A\in \mathcal{I}) \{q_{\mathcal{I},0}(X)-q_{\mathcal{I}'}(X)\}^2|\{O_i\}_{i\in \mathbb{L}_{\ell}^c}]\\+\Mean [\mathbb{I}(A\in \mathcal{I}) \{\widehat{q}_{\mathcal{I}'}(X)-q_{\mathcal{I}',0}(X)\}^2|\{O_i\}_{i\in \mathbb{L}_{\ell}^c}]\\-2\Mean [\mathbb{I}(A\in \mathcal{I}) \{q_{\mathcal{I},0}(X)-q_{\mathcal{I}',0}(X)\}\{\widehat{q}_{\mathcal{I}'}(X)-q_{\mathcal{I}',0}(X)\}|\{O_i\}_{i\in \mathbb{L}_{\ell}^c}].
\end{eqnarray*}
By Cauchy-Schwarz inequality, the last term on the right-hand-side can be lower bounded by
\begin{eqnarray*}
	-\frac{1}{2}\Mean [\mathbb{I}(A\in \mathcal{I}) \{q_{\mathcal{I},0}(X)-q_{\mathcal{I}',0}(X)\}^2|\{O_i\}_{i\in \mathbb{L}_{\ell}^c}]-2\Mean [\mathbb{I}(A\in \mathcal{I}) \{\widehat{q}_{\mathcal{I}'}(X)-q_{\mathcal{I}',0}(X)\}^2|\{O_i\}_{i\in \mathbb{L}_{\ell}^c}].
\end{eqnarray*}
It follows that
\begin{eqnarray*}
	\Mean [\mathbb{I}(A\in \mathcal{I}) \{q_{\mathcal{I},0}(X)-\widehat{q}_{\mathcal{I}'}(X)\}^2|\{O_i\}_{i\in \mathbb{L}_{\ell}^c}]\ge \frac{1}{2}\Mean [\mathbb{I}(A\in \mathcal{I}) \{q_{\mathcal{I},0}(X)-q_{\mathcal{I}',0}(X)\}^2|\{O_i\}_{i\in \mathbb{L}_{\ell}^c}]\\-3\Mean [\mathbb{I}(A\in \mathcal{I}) \{\widehat{q}_{\mathcal{I}'}(X)-q_{\mathcal{I}',0}(X)\}^2|\{O_i\}_{i\in \mathbb{L}_{\ell}^c}],
\end{eqnarray*}
and hence
\begin{eqnarray*}
	\frac{1}{2}\Mean [\mathbb{I}(A\in \mathcal{I}) \{q_{\mathcal{I},0}(X)-q_{\mathcal{I}',0}(X)\}^2|\{O_i\}_{i\in \mathbb{L}_{\ell}^c}]-2\gamma_n+O(1)|\mathcal{I}|(n|\mathcal{I}|)^{-2\beta/(2\beta+p)}\log^8 n\\\le \Mean [\mathbb{I}(A\in \mathcal{I}) \{q_{\mathcal{I},0}(X)-\widehat{q}_{\mathcal{I}}(X)\}^2|\{O_i\}_{i\in \mathbb{L}_{\ell}^c}]
	+3\Mean [\mathbb{I}(A\in \mathcal{I}) \{q_{\mathcal{I}',0}(X)-\widehat{q}_{\mathcal{I}'}(X)\}^2|\{O_i\}_{i\in \mathbb{L}_{\ell}^c}].
\end{eqnarray*}
By Lemma \ref{lemma0}, Lemma \ref{lemma2} and the positivity assumption, the right-hand-side is upper bounded by $O(1)|\mathcal{I}|(n|\mathcal{I}|)^{-2\beta/(p+2\beta)}\log^8 n$ for some universal positive constant $O(1)$, with probability approaching $1$. We obtain with probability approaching $1$ that
\begin{eqnarray*}
	\Mean [\mathbb{I}(A\in \mathcal{I}) \{q_{\mathcal{I}}(X)-q_{\mathcal{I}'}(X)\}^2|\{O_i\}_{i\in \mathbb{L}_{\ell}^c}]=4\gamma_n+O(1)|\mathcal{I}|(n|\mathcal{I}|)^{-2\beta/(2\beta+p)}\log^8 n,
\end{eqnarray*}
uniformly for any $\mathcal{I}$ and $\mathcal{I}'$, or equivalently,
\begin{eqnarray*}
	\Mean \left[\frac{b(\mathcal{I}|X)}{|\mathcal{I}|} \{q_{\mathcal{I}}(X)-q_{\mathcal{I}'}(X)\}^2|\{O_i\}_{i\in \mathbb{L}_{\ell}^c}\right]=\frac{4\gamma_n}{|\mathcal{I}|}+O(1)(n|\mathcal{I}|)^{-2\beta/(2\beta+p)}\log^8 n.
\end{eqnarray*}
By the positivity assumption, we have with probability approaching $1$ that
\begin{eqnarray*}
	\Mean[\{q_{\mathcal{I}}(X)-q_{\mathcal{I}'}(X)\}^2|\{O_i\}_{i\in \mathbb{L}_{\ell}^c}]=O(\gamma_n |\mathcal{I}|^{-1})+O\{(n|\mathcal{I}|)^{-2\beta/(2\beta+p)}\log^8 n\},
\end{eqnarray*}
uniformly for any $\mathcal{I}$ and $\mathcal{I}'$. The proof is hence completed by noting that $\gamma_n$ is at least of the order $O(n^{-2\beta}/(2\beta+p))\log^8 n$. 

\smallskip

\noindent \textit{Part 2}. Consider the bias of the proposed estimator first. Similar to Part 3 of the proof of Theorem \ref{thm1}, the bias is given by $\mathcal{L}^{-1}\sum_{\ell=1}^{\mathcal{L}}V(\pi_{\widehat{\mathcal{D}}^{(\ell)}})-V(\pi)$. By definition,
\begin{eqnarray*}
	V(\pi_{\widehat{\mathcal{D}}^{(\ell)}})-V(\pi)=\sum_{\mathcal{I}\in \widehat{\mathcal{D}}^{(\ell)}} \int_{\mathcal{I}} \Mean Q(X,a) \mathbb{I}(\pi(X)\in \mathcal{I}) \frac{b(a|X)}{b(\mathcal{I}|X)}da-\Mean Q\{X,\pi(X)\}\\
	=\sum_{\mathcal{I}\in \widehat{\mathcal{D}}^{(\ell)}} \int_{\mathcal{I}} \Mean \{Q(X,a)-Q\{X,\pi(X)\}\} \mathbb{I}(\pi(X)\in \mathcal{I}) \frac{b(a|X)}{b(\mathcal{I}|X)}da\\=\sum_{\mathcal{I}'\in \widehat{\mathcal{D}}^{(\ell)}} \Mean\{q_{\mathcal{I},0}(X)-Q\{X,\pi(X)\}\}\mathbb{I}(\pi(X)\in \mathcal{I}).
\end{eqnarray*}
It follows that
\begin{eqnarray}\label{eqn:thm2eq1}
	|V(\pi_{\widehat{\mathcal{D}}^{(\ell)}})-V(\pi)|\le \sup_{\mathcal{I}'\in \widehat{\mathcal{D}}^{(\ell)},a\in \mathcal{I}'} \Mean|Q(X,a)-q_{\mathcal{I}'}(X)|. 
\end{eqnarray} 
For any $\mathcal{I}'\in \widehat{\mathcal{D}}^{(\ell)}$. Consider two separate cases, corresponding to $|\mathcal{I}'|\le \gamma_n^{1/3}$ and $|\mathcal{I}'|> \gamma_n^{1/3}$, respectively. 

In Case 1, it follows from the Lipschitz property of the conditional mean function $Q$ that $|Q(x,a_1)-Q(x,a_2)|\le L \gamma_n^{1/3}$, for any $a_1,a_2\in \mathcal{I}'$ and $x$. By definition, the function $q_{\mathcal{I}'}$ can be represented as $q_{\mathcal{I}'}(x)=\int_{\mathcal{I}'} Q(x,a)\omega(a,x)da$ for some weight function $\omega$ such that $\int_{\mathcal{I}'} \omega(a,x)da=1$. It follows that the right-hand-side of \eqref{eqn:thm2eq1} is upper bounded by $L\gamma_n^{1/3}$. 

In Case 2, for any $a\in \mathcal{I}'$, we can find an interval $\mathcal{I}\subseteq \mathcal{I}'$, $a\in \mathcal{I}$ with length proportional to $\gamma_n^{1/3}$. Using similar arguments in Case 1, we can show that $|Q(x,a)-q_{\mathcal{I},0}(x)|\le L\gamma_n^{1/3}$. By Lemma \ref{lemmacontinuous} and the Cauchy-Schwarz inequality, we have \begin{eqnarray*}\Mean |q_{\mathcal{I},0}(X)-q_{\mathcal{I}',0}(X)|\le \sqrt{\bar{C}\gamma_n^{2/3}}= {\bar{C}}^{1/2}\gamma_n^{1/3},\end{eqnarray*} with probability approaching $1$. It follows that the right-hand-side of 
\eqref{eqn:thm2eq1} is upper bounded by $(L+\sqrt{\bar{C}})\gamma_n^{1/3}$, with probability approaching $1$. 

As such, the bias of the proposed estimator is upper bounded by \begin{eqnarray*}(L+\sqrt{\bar{C}})\gamma_n^{1/3},\end{eqnarray*}
with probability approaching $1$.  

We next consider the standard deviation of our estimator. The proposed estimator is can be represented by $\mathcal{L}^{-1}\sum_{\ell=1}^{\mathcal{L}} \widehat{V}^{\ell}(\pi)$ where $\widehat{V}^{\ell}(\pi)$ is the value estimator constructed based on the samples in $\{O_i\}_{i\in \mathbb{L}_{\ell}}$. Since the propensity score function is known to us, each $\widehat{V}^{\ell}(\pi)$ is unbiased to $V(\pi_{\widehat{\mathcal{D}}^{(\ell)}})$. Under the positivity assumption and the boundedness assumption on $Y$ and $\widehat{q}_{\mathcal{I}}$, the variance of $\widehat{V}^{\ell}(\pi)$ is upper bounded by $|\mathbb{L}_{\ell}|^{-1}\inf_{\mathcal{I}\in \widehat{\mathcal{D}}^{(\ell)}} |\mathcal{I}|^{-1}$. By Lemma \ref{lemma2}, it is upper bounded by $O(n^{-1}\gamma_n^{-1})$. As such, the standard deviation of our estimator is upper bounded by $O(n^{-1}\gamma_n^{-1})$.

As such, the convergence rate is given by $O_p(\gamma_n^{1/3}+n^{-1/2}\gamma_n^{-1/2})$. By setting $\gamma_n=n^{-3/5}$, the rate is given by $O_p(n^{-1/5})$. The proof is hence completed.

\end{document}